\documentclass{ieeetj}
\usepackage{cite}
\usepackage{amsmath,amssymb,amsfonts}
\usepackage{textcomp}
\usepackage{xcolor}
\usepackage{hyperref}
\hypersetup{hidelinks=true}
\usepackage{algorithm,algorithmic}
\def\BibTeX{{\rm B\kern-.05em{\sc i\kern-.025em b}\kern-.08em
    T\kern-.1667em\lower.7ex\hbox{E}\kern-.125emX}}
\AtBeginDocument{\definecolor{tmlcncolor}{cmyk}{0.93,0.59,0.15,0.02}\definecolor{NavyBlue}{RGB}{0,86,125}}
\usepackage[pdftex]{graphicx}
\usepackage{color}
\usepackage{bm}
\usepackage[hang,small,bf]{caption}
\usepackage[subrefformat=parens]{subcaption}
\usepackage{tabularx}

\newtheorem{theorem}{Theorem}
\newtheorem{ass}{Assumption}

\def\OJlogo{\vspace{-4pt}$<$Society logo(s) and publication title will appear here.$>$}
\def\seclogo{\vspace{10pt}$<$Society logo(s) and publication title will appear here.$>$}

\def\authorrefmark#1{\ensuremath{^{\textbf{#1}}}}

\begin{document}
\receiveddate{XX Month, XXXX}
\reviseddate{XX Month, XXXX}
\accepteddate{XX Month, XXXX}
\publisheddate{XX Month, XXXX}
\currentdate{XX Month, XXXX}
\doiinfo{XXXX.2022.1234567}

\markboth{}{Author {et al.}}

\title{Deep Unfolding-based \\ Weighted Averaging \\
for Federated Learning\\ 
in Heterogeneous Environments}

\author{Ayano Nakai-Kasai\authorrefmark{1}, Member, IEEE and Tadashi Wadayama\authorrefmark{1}, Member, IEEE}
\affil{Graduate School of Engineering, Nagoya Institute of Technology, Nagoya, Japan}
\corresp{Corresponding author: Ayano Nakai-Kasai (email: nakai.ayano@nitech.ac.jp).}
\authornote{This work was supported by JSPS KAKENHI Grant-in-Aid for Young Scientists Grant Number JP23K13334 (to A. Nakai-Kasai) 
and for Scientific Research(A) Grant Number JP22H00514 (to T. Wadayama).}

\begin{abstract}
    Federated learning is a collaborative model training method that iterates model updates by multiple clients and aggregation of the updates by a central server.
    Device and statistical heterogeneity of participating clients cause significant performance degradation 
    so that an appropriate aggregation weight should be assigned to each client in the aggregation phase of the server.
    To adjust the aggregation weights, this paper employs deep unfolding, 
    which is known as the parameter tuning method that leverages both 
    learning capability using training data like deep learning and domain knowledge.
    This enables us to directly incorporate the heterogeneity of the environment of interest into the tuning of the aggregation weights.
    The proposed approach can be combined with various federated learning algorithms. 
    The results of numerical experiments indicate that 
    a higher test accuracy for unknown class-balanced data can be obtained with the proposed method 
    than that with conventional heuristic weighting methods.
    The proposed method can handle large-scale learning models with the aid of pretrained models 
    such that it can perform practical real-world tasks.
    Convergence rate of federated learning algorithms with the proposed method is also provided in this paper.
\end{abstract}

\begin{IEEEkeywords}
    Federated learning, deep unfolding, model-based deep learning
\end{IEEEkeywords}


\maketitle

\section{Introduction}
\label{sec:intro}
Federated learning, first proposed in \cite{McMahan}, enables the collaborative distributed training of machine learning models 
using data from many participating devices while keeping the raw data locally.
The general objective of federated learning is to collaboratively obtain a model 
that can provide better predictions for unknown data 
by exploiting a small amount of data from each device while preserving privacy.
It is suitable for processing personal or highly confidential data.
For example, hospitals can collaboratively train medical imaging diagnostic model 
through federated learning using data stored at each hospital 
without leaking data \cite{medical}.
The federated learning system comprises multiple clients 
and a central server that orchestrates the learning process as the service provider.
Clients are devices or institutions with local data, 
such as smartphones, edge Internet-of-Things devices \cite{Yang}, 
and financial and medical organizations \cite{medical}.
A typical federated learning process \cite{Book} iterates 
the following procedure: 
model sharing by the server, 
local model update by each client using its own local data, 
and model aggregation based on the weighted averaging of the clients' updates by the server.

The major challenge in obtaining better models via federated learning is 
\emph{heterogeneity} of the clients and their data.
Device heterogeneity 
and statistical heterogeneity 
exist as the typical factors \cite{Xu,Wu}.
Specifically, clients have different CPUs and memory configurations, and generally operate under various network conditions. 
Moreover, the local data distribution is frequently not independent and identically distributed (non-IID) 
because it comprises client-specific data.
Such scenarios occur in practice depending on the personal preferences or localities of the clients.
Heterogeneity causes insufficient training against comparatively minor data 
and severe performance degradation~\cite{Gafni}.
The method of solving the heterogeneity issue remains debatable, and 
many researchers have attempted to obtain better models for heterogeneous environments.
These attempts are systematically summarized in ~\cite{Book,TLi}.

This problem has been addressed from various perspectives, 
such as improving a client's local update \cite{Si,fedprox,FedNova,SCAFFOLD} and employing sophisticated model updates at the server\cite{Sattler,Xing}.
The simplest approach for a wide range of applications is to control the aggregation weights used in the server's model aaggregation. 
The server aggregates the clients' updates by performing weighted averaging with the aggregation weights assigned to each client.
McMahan et al. proposed the use of aggregation weights that are proportional only to the data quantity \cite{McMahan}.
This weighting strategy seems intuitive and provides empirically stable performance.
However, these aggregation weights are not always optimal in heterogeneous environments \cite{Zhaowireless}. 
This strategy cannot incorporate statistical heterogeneity such as data quality or device heterogeneity such as computational capability.
Some weighting methods use information of training losses \cite{Mohli,TLiarxiv,Zhao}, 
local gradients \cite{FedAdp}, 
training accuracy, and participation frequency \cite{FedFa}.
However, these strategies can be considered as heuristics.
In the presence of complicated heterogeneity as in an actual environment, 
the effect of each factor, such as the data quantity, training loss, and participation frequency, on the learning process is unclear.
Furthermore, the magnitude of these effects is expected to vary considerably depending on the execution environments, 
making it difficult to accurately estimate.
If we can use appropriate aggregation weights that accurately reflect the heterogeneity of the environment of interest, 
then we expect to improve the learning performance.
Therefore, a method that 
can directly adopt the heterogeneity of the environment of interest in the design of aggregation weights should be developed.

In the fields of signal processing and wireless communications, 
methods that incorporate learning capabilities of deep learning into domain knowledge-based algorithms, 
known as deep unfolding or model-based deep learning \cite{Gregor,Monga,Shlezinger}, have been employed to tune certain parameters included in the algorithms. 
This has various applications such as sparse signal recovery \cite{TISTA} and the average consensus problem \cite{consensus}.
In deep unfolding, 
tunable parameters in an algorithm are trained using actual data from the system of interest as training data. 
The resulting parameters, for example, yield an improved performance within a practical number of iterations of the algorithm.
This is achieved by directly incorporating the characteristics of the system of interest into the parameter tuning.

For the assignment of aggregation weights in federated learning, 
it is challenging to properly incorporate heterogeneity of the environment of interest 
with the aforementioned heuristic approaches.
To obtain appropriate aggregation weights that accurately reflect heterogeneity, 
as an alternative and novel approach, 
we can consider incorporating the heterogeneity directly through learning, utilizing actual data and environment of interest.
For this purpose, the technique of deep unfolding is ideally suitable for tuning aggregation weights in federated learning.
The derivation of aggregation weights by learning is expected to provide a better performance than heuristic weighting methods.

In this paper, we employ deep unfolding to numerically optimize the assigned aggregation weights according to the heterogeneity in federated learning.
The optimization process of the aggregation weights based on deep unfolding is performed as a preprocessing phase before federated learning.
The proposed approach can be considered a kind of {\em learning-to-learn} or {\em meta-learning} \cite{Finn,meta}, 
which has not yet been established in the context of federated learning.
The learning procedure of federated learning can be unfolded as a deep network, 
where each layer of the network comprises a process of the federated learning algorithm.
The federated learning process includes neural networks 
so that the unfolded network has a nested network structure.
The aggregation weights included in the weighted averaging step by the server become trainable by using the idea of deep unfolding.
The proposed method using deep unfolding can directly adopt the heterogeneity 
by using target environments as training data 
whereas the data remain locally.
No other methods using the same approach exists.
Therefore, this provides a novel weighting method for federated learning.
Learning the weights to adapt to heterogeneity enables the appropriate assignment, 
and the results may provide insights for developing a novel weighting strategy.
Convergence rate of federated learning algorithm with the proposed method is also provided in this paper.

If appropriate weights depending on the environment of interest can be obtained using the proposed approach, 
it may have a practical impact on challenging tasks, 
such as medical data tasks \cite{medical} that require learning from a very small amount of data 
and extreme classification tasks \cite{Tagami}, in which non-IIDness is likely to be strong.
Furthermore, we expect it to improve the performance of practical tasks using large-scale models, 
which have attracted considerable interest in recent years \cite{transformer}.
The proposed weight optimization using deep unfolding can incorporate large-scale classification models such as a vision transformer (ViT) \cite{vit} 
with the aid of pretrained models.
The proposed approach can provide models that can perform practical image classification tasks in federated learning.
This is supported by the experimental results in this paper obtained using real data.


\section{Related Work}
\label{sec:related}
Non-IID data distribution is an inseparable problem in federated learning 
dealing with clients' personal data.
The factors of non-IIDness are categorized in \cite{Book,Zhu}.
The effect of non-IIDness on the test performance was experimentally demonstrated in \cite{Zhaoexp}, 
which indicated that the test accuracy may decreases by 50\% in non-IID environments.
Many studies have attempted to improve the performance with non-IID data 
by modifying local update rules \cite{Si,fedprox,FedNova,SCAFFOLD}, by using clustering \cite{Sattler,Briggs} or graph learning \cite{Xing}, 
or by combining with reinforcement learning \cite{Kang}.
These approaches are orthogonal and can be combined with the proposed weighting method.
Another context of federated learning aims to obtain a model that provides better predictions for each client's non-IID data.
It is called personalized federated learning \cite{Tan,Huang,Jiang,Fallah,Hong}.

The convergence rate of the first federated learning algorithm, federated averaging (FedAvg) \cite{McMahan}, 
is discussed for both IID \cite{Stich} and non-IID datasets \cite{Li}.
Other convergence rates are summarized in \cite{Book}.
In most cases, the rate is proportional to $1/\sqrt{T}$, where $T$ is the number of iteration rounds of an algorithm.

Many heuristic weighting methods exists for solving non-IID and heterogeneity problems \cite{FOCUS,Kim,Hong}.
If clients share part of the data with the central server, 
the weighting method proposed in \cite{Zhaowireless} can be applied.
Other weighting methods consider fairness \cite{Mohli,TLiarxiv,Zhao}.
The fairness in federated learning is defined as follows: 
certain model parameters are considered fair solutions for the predefined objective 
if the accuracy distribution of the model parameters is uniform \cite{TLiarxiv,Zhao}.
These methods additionally use training losses of clients.
Zhao and Joshi \cite{Zhao} recently derived a dynamic reweighting (DR) strategy that imposes penalties on clients depending on training loss.
A weighting method that considers the communication efficiency was proposed for more general scenarios \cite{Chen1}. 
However, it requires calculations for all potential clients and is unsuitable for cross-device scenarios involving hundreds of clients.
The method \cite{Wang} uses the Sharpley value of each client at an additional cost.
Wu and Wang \cite{FedAdp} proposed a scalable weighting method called federated adaptive weighting (FedAdp) 
based on the correlation between the local and global gradients.
Huang et al. \cite{FedFa} proposed fairness and accuracy in horizontal federated learning (FedFa) 
that uses training accuracy and communication frequency as measures of a client's contribution.
FedAdp and FedFa can incorporate non-IIDness and device heterogeneity.

Client selection is related to the weighting strategies.
The original FedAvg \cite{McMahan} and other methods \cite{Li,Cho,Chen3,Ribero,Chen2} 
employ client selection, 
where the central server samples clients expected to make a high contribution to the learning process more frequently as participants.
Adopting client selection can lead to the learning involving a large number of potential clients.
In practical scenarios, however, the selected clients cannot always participate in the learning process 
owing to the network conditions and load status of the devices \cite{Xu,MATCHA}.
In such cases, federated learning methods can be made even more effective and scalable 
by combining client selection and weighting strategies.


In deep unfolding or model-based deep learning \cite{Gregor,Monga,Shlezinger}, 
the objective is to improve the performance of the target algorithm or model 
by tuning certain embedded parameters.
The approach that mitigates the constraints on the structure of the models and algorithms is generally called meta-learning \cite{Finn,meta}.
Ren et al. and Shu et al. proposed sample weighting methods to manage biased training data using a meta-learning approach \cite{Ren,Shu}.
Deep unfolding with a nested network structure was employed in \cite{Nagahama} 
for the context of graph signal processing.

\section{Preliminaries}
\subsection{Notation}
In the reainder of this paper, we use the following notations:
The superscript $(\cdot)^{\mathrm{T}}$ denotes a transpose operation.
The zero vector and the vector whose elements are all $1$ are represented by $\bm{0}$ and $\bm{1}$, respectively.
The Euclidean ($\ell_2$) norm is denoted by $\|\cdot\|$, and $\ell_1$ norm is denoted by $\|\cdot\|_1$.
The expectation operator is $\mathbb{E}[\cdot]$.

\subsection{Federated Learning}
Let us consider a system comprising $K$ clients and a single central server. 
Each client has its own local dataset.
The number of local data samples for client $k \ (k=1,\ldots,K)$ is $N_k$, and the total number is $N=\sum_{k=1}^K N_k$.
The clients and the central server are assumed to have the same objective 
of finding a set of learning model parameters $\bm{w}_\mathrm{opt}\in\mathbb{R}^d$ that minimizes the objective function $f:\mathbb{R}^d\to\mathbb{R}$ 
without sharing the local data, that is, 
\begin{equation}
  \bm{w}_\mathrm{opt} = \mathrm{arg}\min_{\bm{w}\in\mathbb{R}^d} f(\bm{w}).
\end{equation}
A typical objective function has the following finite-sum form: 
\begin{align}
  f(\bm{w}) = \sum_{k=1}^K \theta_k f_k(\bm{w}),
  \label{eq:objective}
\end{align}
where $\theta_k\in\mathbb{R}$ is a nonnegative weight for client $k$ that satisfies $\sum_{k=1}^K \theta_k=1$, 
and $f_k:\mathbb{R}^d\to\mathbb{R}$ is the client $k$'s local objective function 
composed of loss function values.

In typical federated learning algorithms, 
clients and server iterate (1) \emph{client update}: 
each client performs local updates of its model using its own local data, 
and (2) \emph{aggregation}: 
the server aggregates the clients' models obtained by the client update.
The specific learning processes of some federated learning algorithms are briefly reviewed in the following subsections.

\subsection{FedAvg}
FedAvg \cite{McMahan} is the best-known federated learning algorithm.
FedAvg iterates client selection, 
parallel client updates of the model parameters based on vanilla stochastic gradient descent (SGD) at the selected clients, 
and model aggregation at the server via weighted averaging of the model parameters, 
to collaboratively find the optimal model parameter.

Let $\ell(\bm{x}_k,\bm{y}_k;\bm{w})$ be client $k$'s loss yielded by the model parameter $\bm{w}$ 
on its local dataset with inputs $\bm{x}_k$ and corresponding outputs $\bm{y}_k$, 
and $\nabla\ell(\bm{x}_k,\bm{y}_k;\bm{w})$ be its gradient.
The procedure of FedAvg and the local client update based on vanilla SGD are summarized in Algorithms~\ref{alg:fedavg} and \ref{alg:sgd}, respectively, 
where $T$ is the number of iteration rounds, 
$E$ is the number of local epochs, and $\eta$ is the learning rate.
Note that the detailed processes for client selection are omitted 
because they are beyond the focus of this paper.

The weight 
\begin{align}
  \theta_k^\mathrm{Avg} = \frac{N_k}{N}  
  \label{eq:thetafedavg}
\end{align} 
is used for the aggregation step at the server (line 5 of Algorithm~\ref{alg:fedavg}): 
\begin{equation}
  \bm{w}^{(t+1)} = \sum_{k=1}^K \frac{N_k}{N} \bm{w}_k^{(t)} = \sum_{k=1}^K \theta_k^{\mathrm{Avg}} \bm{w}_k^{(t)}.
\end{equation}
This weighting rule can be regarded as reflecting the concept that 
a client with more data contributes more to learning 
if the data distributions are IID and the clients are in a homogeneous environment.
\begin{algorithm}[tb]
	\caption{FedAvg \cite{McMahan}}
	\label{alg:fedavg}
	\begin{algorithmic}[1]
    \STATE Server: initialize model parameter $\bm{w}^{(0)}$
		\FOR{$t=0,\ldots,T-1$}
			\STATE Server: select $K$ clients and share $\bm{w}^{(t)}$
			\STATE Each client $k$: $\bm{w}^{(t)}_k=$ {\sf ClientUpdate}$(k,\bm{w}^{(t)})$
      \STATE Server: $\bm{w}^{(t+1)}=\sum_{k=1}^K \frac{N_k}{N}\bm{w}^{(t)}_k$
		\ENDFOR
    \RETURN $\bm{w}^{(T)}$
	\end{algorithmic}
\end{algorithm}
\begin{algorithm}[tb]
	\caption{Client Update based on Vanilla SGD}
	\label{alg:sgd}
	\begin{algorithmic}[1]
		\STATE {\sf ClientUpdate}$(k,\bm{w})$:
      \FOR{$e=1,\ldots,E$}
        \STATE Construct the set $\mathcal{B}$ of local minibatches of $(\bm{x}_k,\bm{y}_k)$
        \FOR{$(\bm{x}_b,\bm{y}_b)\in\mathcal{B}$}
          \STATE $\bm{w} = \bm{w}-\eta\nabla\ell(\bm{x}_b,\bm{y}_b;\bm{w})$
        \ENDFOR
      \ENDFOR
      \RETURN $\bm{w}$
	\end{algorithmic}
\end{algorithm}

\subsection{General Formulation}
\label{subsec:general}
The variants of federated learning algorithms including FedAvg can be summarized as 
the following formulations \cite{FedNova}: 
\begin{align}
  \bm{w}_k^{(t)} &= \bm{w}^{(t)} - \tau_{\mathrm{eff}}\sum_{\tau=0}^{\tau_k-1}\eta\frac{a_{k,\tau}\nabla\ell(\bm{w}_{k,t}^\tau)}{\|\bm{a}_k\|_1}, \ (k=1,\ldots,K),\\
  \bm{w}^{(t+1)} &= \sum_{k=1}^K \tilde{\theta}_k\bm{w}_k^{(t)}, 
\end{align}
where $\nabla\ell(\bm{w}_{k,t}^\tau)\in\mathbb{R}^d$ and $\bm{w}_{k,t}^\tau\in\mathbb{R}^d$ are the gradient of loss and the model parameter of client $k$ in $\tau$th client update in round $t$, 
$\tau_k\geq0$ is the number of client updates for client $k$ in each round, and $\bm{w}_{k,t}^0=\bm{w}^{(t)}$.
The vector $\bm{a}_k=[a_{k,0}, \ldots, a_{k,{\tau_k-1}}]^\mathrm{T}\in\mathbb{R}^{\tau_k}$ is determined depending on an algorithm.
The parameter $\tilde{\theta}_k$ is the nonnegative aggregation weight required to satisfy $\sum_{k=1}^K\tilde{\theta}_k=1$.
The constant $\tau_{\mathrm{eff}}$ is the effective number of client updates.
These equations can be alternatively expressed as 
\begin{equation}
    \bm{w}^{(t+1)} = \bm{w}^{(t)}-\tau_{\mathrm{eff}}\sum_{k=1}^K \tilde{\theta}_k\eta \tilde{\bm{d}}_k^{(t)}, 
    \label{eq:general}
\end{equation}
where 
\begin{equation}
    \tilde{\bm{d}}_k^{(t)}=\frac{\tilde{\bm{G}}_k^{(t)}\bm{a}_k}{\|\bm{a}_k\|_1}
    \label{eq:dkt}
\end{equation}
and $\tilde{\bm{G}}_k^{(t)}=[\nabla\ell(\bm{w}_{k,t}^0), \ldots, \nabla\ell(\bm{w}_{k,t}^{\tau_k-1})]\in\mathbb{R}^{d\times \tau_k}$.

For example, we can reconstruct FedAvg \cite{McMahan} by setting 
\[
  \bm{a}_k=[1,\ldots,1]^\mathrm{T}, \ 
  \tau_{\mathrm{eff}}=\sum_{k=1}^K \frac{N_k}{N}\tau_k, \
  \tilde{\theta}_k=\frac{\frac{N_k}{N}\tau_k}{\sum_{l=1}^K \frac{N_l}{N}\tau_l}.
\]
For FedProx \cite{fedprox}, 
\begin{align}
  \bm{a}_k&=[(1-\alpha)^{\tau_k-1}, (1-\alpha)^{\tau_k-2}, \ldots, (1-\alpha), 1]^\mathrm{T}, \nonumber \\
  \tau_{\mathrm{eff}}&=\frac{1}{\alpha} \sum_{k=1}^K \frac{N_k}{N}(1-(1-\alpha)^{\tau_k}), \nonumber \\
  \tilde{\theta}_k&=\frac{\frac{N_k}{N}(1-(1-\alpha)^{\tau_k})}{\sum_{l=1}^K\frac{N_l}{N}(1-(1-\alpha)^{\tau_l})}, \nonumber
\end{align}
where $\alpha=\eta\mu$, and $\mu\geq0$ is a hyperparameter included in the algorithm.

In addition, formulation \eqref{eq:general} can include 
not only vanilla SGD but also various types of local updates 
with learning rate scheduling and momentum \cite{FedNova}.

\subsection{FedNova}
FedNova \cite{FedNova} can incorporate a time-varying number of client updates caused by device performance or communication restrictions  
by introducing round-dependent parameters $\tau_k^{(t)}\geq0$ and $\bm{a}_k^{(t)}\in\mathbb{R}^{\tau_k^{(t)}}$.
The update equation of FedNova is given by 
\begin{equation}
    \bm{w}^{(t+1)} = \bm{w}^{(t)}-\tau_{\mathrm{eff}}^{(t)}\sum_{k=1}^K \frac{N_k}{N}\eta \bm{d}_k^{(t)}, 
    \label{eq:generalt}
\end{equation}
where 
\begin{equation}
    \bm{d}_k^{(t)}=\frac{\bm{G}_k^{(t)}\bm{a}_k^{(t)}}{\|\bm{a}_k^{(t)}\|_1},
    \label{eq:dktnew}
\end{equation}
$\bm{G}_k^{(t)}=[\nabla\ell(\bm{w}_{k,t}^0), \ldots, \nabla\ell(\bm{w}_{k,t}^{\tau_k^{(t)}-1})]\in\mathbb{R}^{d\times \tau_k^{(t)}}$, 
and $\bm{a}_k^{(t)}=[a_{k,0}^{(t)}, \ldots, a_{k,\tau_k^{(t)}-1}^{(t)}]^\mathrm{T}\in\mathbb{R}^{\tau_k^{(t)}}$.
In FedNova, the weight 
\begin{equation}
  \tilde{\theta}_k^\mathrm{Nova} = \theta_k^\mathrm{Avg} = \frac{N_k}{N}
  \label{eq:thetafednova}
\end{equation}
is used for the weighted averaging process.
One can set $\tau_{\mathrm{eff}}^{(t)}=\sum_{k=1}^K N_k\tau_k^{(t)}/N$ 
if one considers consistency with FedAvg.

\subsection{Related Weighting Strategies}
The data quantity-based weights $\theta_k^{\mathrm{Avg}}$ employed in FedAvg \eqref{eq:thetafedavg} and FedNova \eqref{eq:thetafednova} tend to emphasize the information from 
the client with a large amount of data. 
However, \cite{Zhaowireless} clarified the following: 
such weights are not suitable in heterogeneous environments from the perspective of the divergence of the model parameters 
and alternative weights should be introduced to account for heterogeneity.
In heterogeneous environments, a client with more data does not necessarily contribute significantly to performance.
Specifically, when there are clients having few but unique data that are not included in other clients, 
emphasizing the information from such clients may provide a model with high accuracy for unknown data distribution.
Clients with more data but low computational capabilities and/or poor network conditions 
should not necessarily be emphasized. 
Therefore, the heterogeneity of the clients must be reflected in the selection of the weights.

The recently proposed DR method \cite{Zhao} aims at a fair federated learning 
using the concept that higher weights are assigned to clients with high losses in each round. 
This aids clients with poor learning results.
The weight at $t$th round is given by 
\begin{align}
  \theta_k^{(t)\mathrm{DR}}=\frac{\frac{N_k}{N}(\ell(\bm{x}_k,\bm{y}_k;\bm{w}^{(t)}))^{q+1}}{\sum_{l=1}^K \frac{N_l}{N}(\ell(\bm{x}_l,\bm{y}_l;\bm{w}^{(t)}))^{q+1}}, 
  \label{eq:thetadr}
\end{align}
where $q\ (\geq0)$ is a hyperparameter, and $\bm{w}^{(t)}$ is the model parameter in the previous round.
The authors of \cite{Zhao} proposed an algorithm applying the weights to FedAvg called DR-FedAvg.
The hyperparameter $q$ must be manually selected depending on the heterogeneity to obtain high accuracy.

FedAdp \cite{FedAdp} is a weighting method that uses the correlations between local and global gradients.
This method assigns higher weights to clients with gradients that are close to the global gradient 
to correct the direction of the gradient.
In $t$th round, each client performs $\tau_k^{(t)}$ iterations of the vanilla SGD update, 
and the accumulated local gradient in this step is given by $\bm{g}_k^{(t)}=\bm{G}_k^{(t)}\bm{1}$.
The global gradient is calculated as $\bm{g}^{(t)}=\sum_{k=1}^KN_k\bm{g}_k^{(t)}/N$.
The correlation between the local gradient of each client and global gradient is measured by the angle $\phi_k^{(t)}$ defined by 
\begin{equation}
  \phi_k^{(t)} = \mathrm{arc}\cos\frac{(\bm{g}^{(t)})^\mathrm{T}\bm{g}_k^{(t)}}{\|\bm{g}^{(t)}\|\|\bm{g}_k^{(t)}\|}.
\end{equation}
A larger angle implies a smaller contribution from the client to the global update.
The following smoothed angle is alternatively used to reduce the fluctuations in the angle: 
\begin{equation}
  \tilde{\phi}_k^{(t)} = \begin{cases}
    \phi_k^{(t)}, & \mathrm{if}~t=0, \\
    \frac{t}{t+1}\tilde{\phi}_k^{(t-1)}+\frac{1}{t+1}\phi_k^{(t)}, & \mathrm{otherwise.}
  \end{cases}
\end{equation}
Finally, the smoothed angle is mapped to the following weight: 
\begin{equation}
  \theta_k^{(t)\mathrm{Adp}} = \frac{N_k\exp{(h(\tilde{\phi}_k^{(t)}))}}{\sum_{l=1}^K N_l\exp{(h(\tilde{\phi}_l^{(t)}))}}, 
  \label{eq:fedadp}
\end{equation}
where $h(\phi)=\beta(1-\exp{(-\exp{(-\beta(\phi-1))})})$ and $\beta>0$ is a hyperparameter.

Huang et al. proposed a federated algorithm called FedFa \cite{FedFa} that includes a novel weighting method.
FedFa introduces the accuracy during the training process and the participation frequency of each client into the weight.
A larger weight is assigned to a client with a highly accurate model and a higher potential for participation.
The measure in terms of the training accuracy of each client $k$ is calculated as 
\begin{equation}
  \bar{M}_k^{(t)} = \frac{-\log_2(\bar{L}_k^{(t)})}{-\sum_{l=1}^K \log_2(\bar{L}_l^{(t)})}, 
  \label{eq:acc}
\end{equation}
where $L_l^{(t)}$ is the accuracy calculated using the training data of client $l$ in $t$th round 
and $\bar{L}_k^{(t)}=L_k^{(t)}/\sum_{l=1}^K L_l^{(t)}$.
The measure in terms of the participation frequency of each client $k$ is calculated as 
\begin{equation}
  \bar{E}_k^{(t)} = \frac{-\log_2(1-\bar{F}_k^{(t)})}{-\sum_{l=1}^K \log_2(1-\bar{F}_l^{(t)})}, 
  \label{eq:freq}
\end{equation}
where $F_l^{(t)}$ is the number of training participations of client $l$ up to $t$th round, 
and $\bar{F}_k^{(t)}=F_k^{(t)}/\sum_{l=1}^K F_l^{(t)}$. 
If $\bar{L}_k^{(t)}=0$ or $1-\bar{F}_k^{(t)}=0$, the term is replaced with a very small positive constant to prevent divergence.
From \eqref{eq:acc} and \eqref{eq:freq}, the weight of FedFa is given by 
\begin{equation}
  \theta_k^{(t)\mathrm{Fa}}=\gamma\bar{M}_k^{(t)}+(1-\gamma)\bar{E}_k^{(t)}, 
  \label{eq:fedfa}
\end{equation}
where $\gamma$ is a hyperparameter satisfying $0\leq\gamma\leq1$.

\section{Deep Unfolding-based Weighed Averaging}
\label{sec:proposed}
This section provides the proposed weight optimization approach 
and a theoretical result in terms of federated learning algorithm with the proposed method.

\subsection{Deep Unfolding}
Deep unfolding was first proposed in \cite{Gregor} to improve the performance of sparse signal reconstruction.
An iterative algorithm can be unfolded as a deep network 
with each layer comprising a single step of the algorithm.
This enables us to tune some of the parameters included in the algorithm  
by applying well-known deep learning techniques such as SGD and backpropagation to the unfolded network.
Several excellent survey papers have been published \cite{Monga,Shlezinger}.

The procedure for deep unfolding is described as follows:
First, one identifies an iterative optimization algorithm of interest 
and determines the number of iterations.
Second, the iterative algorithm is unfolded into a network, and 
some trainable parameters are embedded in each layer.
Third, one prepares training data and sets a loss function for the learning process.
Finally, the trainable parameters are tuned with the training data using deep learning techniques 
such that the iterative algorithm achieves the desired performance.


\subsection{Deep Unfolding-based Weighed Averaging}
\begin{figure}[tb]
  \centering
      \includegraphics[width=\columnwidth]{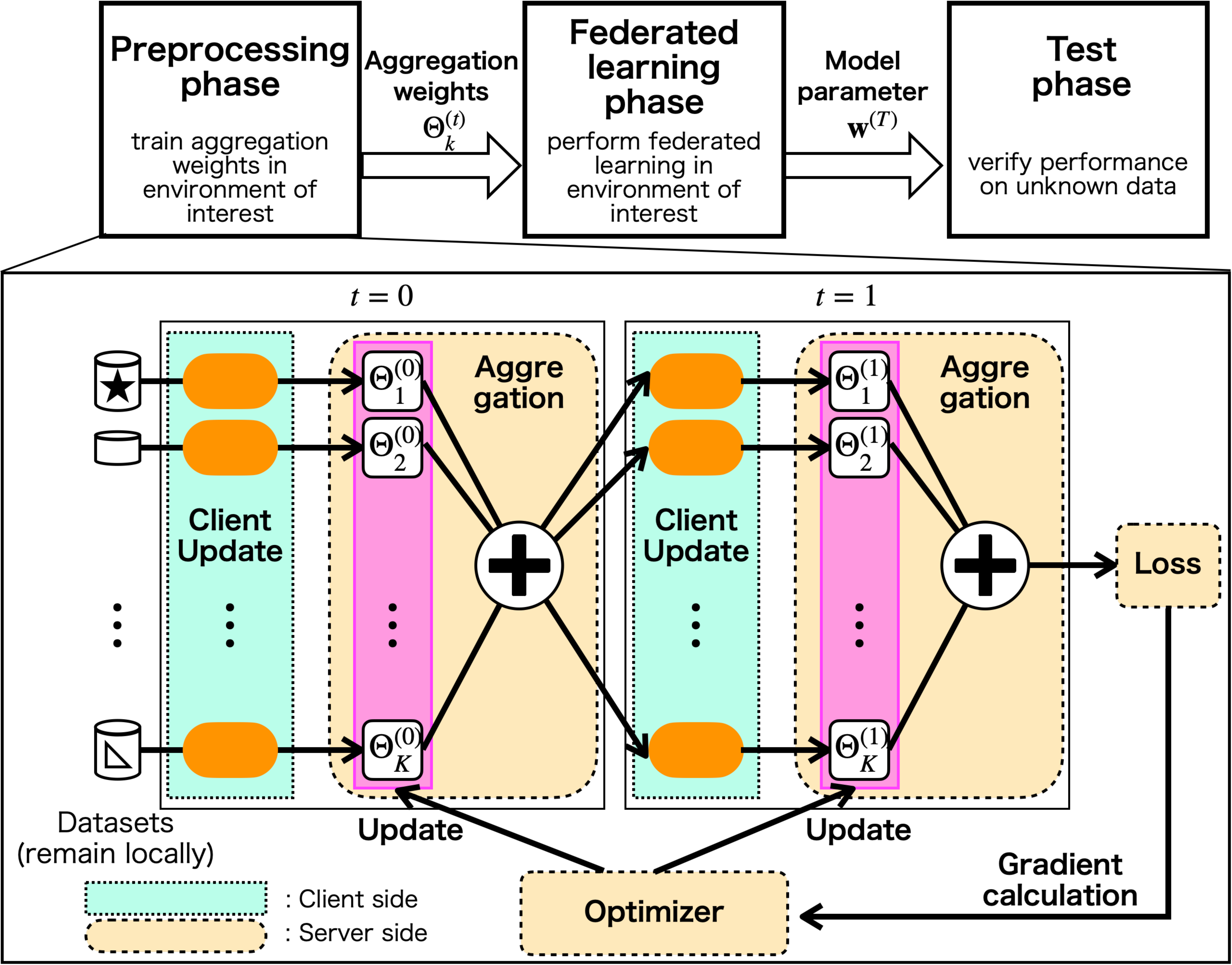}
      \caption{Training process of deep unfolding-based weights $(T=2)$.}
      \label{fig:training}
\end{figure}
Using the concept of deep unfolding, this paper aims to obtain appropriate aggregation weights for federated learning that can directly incorporate heterogeneity. 
This concept is depicted in Figure~\ref{fig:training}.
The training of aggregation weights using deep unfolding is performed as a preprocessing phase. 
Subsequently, federated learning phase is executed using the optimized weights.

The learning procedure for federated learning itself can be unfolded into a deep network, 
where each layer represents a process in a single round of the federated learning.
In this case, the trainable parameters are the aggregation weights $\{\Theta_k^{(t)}\}$ in each round $t$.
In other words, we replace the weighted averaging procedure at the server in Algorithm~\ref{alg:fedavg} as 
\begin{align}
  \bm{w}^{(t+1)}=\sum_{k=1}^K \Theta_k^{(t)}\bm{w}_k^{(t)},
\end{align}
and optimize the weights $\{\Theta_k^{(t)}\}_{t=0,k=1}^{T-1,K}$ using deep learning techniques.

We consider a system in which the server acts as the executor of the preprocessing phase.
In $m$th training step $(m=0,\ldots,M-1)$, the following procedures are iterated:
\begin{enumerate}
  \item Clients and server perform $T$ round federated learning using weights $\{\Theta_k^{(t)}\}_{t=0,k=1}^{T-1,K}$, 
  \item Server calculates loss, 
  \item Server update the weights using optimization method.
\end{enumerate} 

This concept can be applied to various federated learning algorithms such as FedProx \cite{fedprox} and adaptive federated optimization methods \cite{Si}, 
and can adopt various operations.
The following subsections describe the detailed learning procedure for a specific case.

Note that 
the data for each client can be stored locally during the proposed training process.
The clients can provide a part of or all the data for the training process.
However, they can avoid sharing raw data 
by only exchanging model parameters and losses, 
and performing backpropagation in terms of weights via communication.
In other words, heterogeneity such as device states, network conditions, and statistical properties of the data 
can be included in the proposed training process 
while also protecting privacy.

Another point to note is that the proposed method incurs the overhead of preparing a dataset 
and performing training as the preprocessing phase.
A tradeoff may occur between the performance improvement by learning the aggregation weights and the computational load of the overhead part, 
although privacy is protected.
As long as the client is cooperative during the preprocessing 
and can provide some of or all the data for the preprocessing, 
a practically useful approach is 
to obtain the weights 
that are appropriate for the execution environment expected in an actual federated learning process in advance using the proposed method.
The proposed approach enables to obtain appropriate weights before the actual federated learning process.
Moreover, learning the appropriate weights using the proposed approach 
has potential to provide insight into novel adaptive weighting rules.
By incorporating the tendency of the learned weights obtained through the proposed approach, 
more robust and scalable weighting rules can be developed.

\subsection{Example: DUW-FedAvg}
\label{subsec:duwfedavg}
\begin{algorithm}[tb]
	\caption{Training of Deep Unfolding-based Weights for FedAvg}
	\label{alg:prop}
	\begin{algorithmic}[1]
    \FOR{$m=0,\ldots,M-1$}
      \STATE Server: initialize model parameter $\bm{w}^{(0)}$
      \FOR{$t=0,\ldots,T-1$}
        \STATE Server: share model parameter $\bm{w}^{(t)}$
        \STATE Each client $k$: $\bm{w}^{(t)}_k=$ {\sf ClientUpdate}$(k,\bm{w}^{(t)})$, calculate loss $\ell_k^{(t)}$
        \STATE Server: $\bm{w}^{(t+1)}=\sum_{k=1}^K \Theta_k^{(t)}\bm{w}^{(t)}_k$
      \ENDFOR
      \STATE Server: $\ell^{[m]}=\sum_{t=0}^{T-1}\sum_{k=1}^K \ell_k^{(t)}$
      \STATE Server: Update $\{\Theta_k^{(t)}\}$ by optimization method with $\ell^{[m]}$
    \ENDFOR
    \RETURN $\{\Theta_k^{(t)}\}_{t=0,k=1}^{T-1,K}$
	\end{algorithmic}
\end{algorithm}
In this subsection, we present the proposed training process when it is applied to FedAvg.
We assume that the system consists of reliable and stateful devices, 
for example, cross-silo settings \cite{Book}.
Client selection is not assumed to be included.
The proposed method aims to achieve a high accuracy on class-balanced test data not included in training data.

The training procedure is summarized in Algorithm~\ref{alg:prop}.
At $m$th $(m=0,\ldots,M-1)$ learning iteration, $T$ rounds of FedAvg are executed with fixed weights $\{\Theta_k^{(t)}\} \ (t=0,\ldots,T-1)$.
Losses are accumulated during the process and sent to the server.
The weights are updated with the losses using a standard optimization method used in deep learning such as Adam.
FedAvg with the optimized weights is then performed after this preprocessing phase.
We refer to the FedAvg with the learned Deep Unfolding-based Weights (DUW) as DUW-FedAvg.

\subsection{Example: DUW-FedNova}
This subsection describes a case in which the proposed method is applied to another algorithm, FedNova \cite{FedNova}.
As in the previous subsection, client selection is not assumed to be included, 
and the proposed method aims to achieve a high accuracy on class-balanced test data not included in training data.

The training procedure and local client updates are summarized in Algorithms~\ref{alg:prop2} and \ref{alg:nova}, respectively.
Most training procedures are the same as those used for FedAvg 
but only the client update is different. 
We refer to the FedNova with the learned DUW as DUW-FedNova.
This formulation includes DUW-FedAvg in Sect.~\ref{sec:proposed}-\ref{subsec:duwfedavg} as a special case.
\begin{algorithm}[tb]
	\caption{Training of Deep Unfolding-based Weight for FedNova}
	\label{alg:prop2}
	\begin{algorithmic}[1]
    \FOR{$m=0,\ldots,M-1$}
      \STATE Server: initialize model parameter $\bm{w}^{(0)}$
      \FOR{$t=0,\ldots,T-1$}
        \STATE Server: share model parameter $\bm{w}^{(t)}$
        \STATE Each client $k$: $\bm{w}^{(t)}_k=$ {\sf ClientUpdateNova}$(k,t,\bm{w}^{(t)})$, calculate loss $\ell_k^{(t)}$
        \STATE Server: $\bm{w}^{(t+1)}=\sum_{k=1}^K \Theta_k^{(t)}\bm{w}^{(t)}_k$
      \ENDFOR
      \STATE Server: $\ell^{[m]}=\sum_{t=0}^{T-1}\sum_{k=1}^K \ell_k^{(t)}$
      \STATE Server: Update $\{\Theta_k^{(t)}\}$ by optimization method with $\ell^{[m]}$
    \ENDFOR
    \RETURN $\{\Theta_k^{(t)}\}_{t=0,k=1}^{T-1,K}$
	\end{algorithmic}
\end{algorithm}
\begin{algorithm}[tb]
	\caption{Client Update for FedNova \cite{FedNova}}
	\label{alg:nova}
	\begin{algorithmic}[1]
		\STATE {\sf ClientUpdateNova}$(k,t,\bm{w})$:
      \FOR{$\tau=0,\ldots,\tau_k^{(t)}$}
        \STATE $\bm{w} = \bm{w}-\eta\tau_{\mathrm{eff}}^{(t)}\frac{a_{k,\tau}^{(t)}\nabla\ell(\bm{w})}{\|\bm{a}_k^{(t)}\|_1}$
      \ENDFOR
      \RETURN $\bm{w}$
	\end{algorithmic}
\end{algorithm}

\subsection{Convergence Analysis}
This section provides a convergence analysis of the federated algorithms presented in Sect.~\ref{sec:proposed}-\ref{subsec:general} 
using the proposed deep unfolding-based weights.
A federated algorithm converges in the sense of Theorem~\ref{theo:convergence} 
if the algorithm is expressed using the following update equations: 
\begin{align}
  \bm{w}_k^{(t+1)} &= \bm{w}^{(t)}-\tau_{\mathrm{eff}}^{(t)} \eta \bm{d}_k^{(t)}, \ (k=1,\ldots,K), \label{eq:genlocal}\\
  \bm{w}^{(t+1)} &= \sum_{k=1}^K \Theta_k^{(t)}\bm{w}_k^{(t)}, \label{eq:genglobal}
\end{align}
and if the weights $\{\Theta_k^{(t)}\}$ are appropriately trained.

A general objective function 
\begin{equation}
  f(\bm{w}) = \sum_{k=1}^K \theta_k f_k(\bm{w})
\end{equation}
is considered in this subsection, 
where each $\theta_k$ is nonnegative, and $\sum_{k=1}^K\theta_k=1$.
Assumptions \ref{ass:smooth}--\ref{ass:dis} below are commonly used for 
analyses of SGD \cite{Stich,Leon} and federated algorithms \cite{Book,Li,FedNova,SCAFFOLD,Si,MATCHA}, 
and they are also adopted in this paper.
\begin{ass}
  \label{ass:smooth}
  Each local function $f_k(\cdot)$ is $L-$Lipschitz smooth, i.e., for any parameter $\bm{u},\bm{v}$ and $k=1,\ldots,K$, 
  \begin{equation}
    \|\nabla f_k(\bm{u})-\nabla f_k(\bm{v})\|\leq L\|\bm{u}-\bm{v}\|.
  \end{equation}
\end{ass}
\begin{ass}
  \label{ass:sigma}
  For each client, the stochastic gradient is an unbiased estimator of the local gradient, i.e., 
  for any parameter $\bm{w}_k$ for clients $k$ and $k=1,\ldots,K$, 
  \begin{equation}
    \mathbb{E} [\nabla\ell(\bm{w}_k|b)]=\nabla f_k(\bm{w}_k), 
  \end{equation}
  where $b$ denotes a stochastic minibatch.
  For all clients, the variance of the stochastic gradient is bounded by a constant $\sigma^2\geq0$, i.e., 
  for any parameter $\bm{w}_k$ for clients $k$ and $k=1,\ldots,K$, 
  \begin{equation}
    \mathbb{E}\left[\|\nabla\ell(\bm{w}_k|b)-\nabla f_k(\bm{w}_k)\|^2\right]\leq\sigma^2.
  \end{equation}
\end{ass}
\begin{ass}
  \label{ass:G}
  For all clients, the expected squared norm of the stochastic gradient is bounded by a constant $G^2\geq0$, i.e., 
  for any parameter $\bm{w}_k$ for clients $k$ and $k=1,\ldots,K$, 
  \begin{equation}
    \mathbb{E}\left[\|\nabla\ell(\bm{w}_k|b)\|^2\right]\leq G^2,
  \end{equation}
  where $b$ denotes a stochastic minibatch.
\end{ass}
\begin{ass}
  \label{ass:dis}
  The dissimilarity between the local and global objective functions 
  is bounded with the constants $\beta^2\geq1$ and $\kappa^2\geq0$ 
  for any nonnegative weights $\{\omega_k\}$ satisfying $\sum_{k=1}^K\omega_k=1$, i.e., 
  for any parameter $\bm{w}$, 
  \begin{equation}
    \sum_{k=1}^K \omega_k\|\nabla f_k(\bm{w})\|^2\leq\beta^2\left\|\sum_{k=1}^K\omega_k\nabla f_k(\bm{w})\right\|^2+\kappa^2.
  \end{equation}
\end{ass}


Under these assumptions, Theorem~\ref{theo:convergence} establishes convergence rate to 
a stationary point of the objective function $f(\bm{w})=\sum_{k=1}^K\theta_k f_k(\bm{w})$, 
where $\theta_k=\sum_{t=0}^{T-1}\Theta_k^{(t)}/T$.
\begin{theorem}
  \label{theo:convergence}
  Let an objective function be $f(\bm{w})=\sum_{k=1}^K\theta_k f_k(\bm{w})$, 
  where $\theta_k=\sum_{t=0}^{T-1}\Theta_k^{(t)}/T$.
  Assume a sufficiently small learning rate $\eta=\sqrt{K/(\tilde{\tau}T)}$, $\bar{\tau}^{(t)}=\sum_{k=1}^K\tau_k^{(t)}/K$, and $\tilde{\tau}=\sum_{t=0}^{T-1}\bar{\tau}^{(t)}/T$.
  Under Assumptions \ref{ass:smooth}--\ref{ass:dis}, 
  the algorithm described by the update equations \eqref{eq:genlocal} and \eqref{eq:genglobal} converges to a stationary point of the objective function $f$ 
  at the rate of $\mathcal{O}(1/\sqrt{T})+\mathcal{O}(T^\delta)$, 
  where $\mathcal{O}(T^\delta)$ represents the order due to the variance of learned weights $\{\Theta_k^{(t)}\}$ of $T$ rounds for $k=1,\ldots,K$ obtained using deep unfolding.
\end{theorem}
A detailed derivation is provided in Appendix and is primarily based on the convergence results in \cite{FedNova}.

Theorem~\ref{theo:convergence} implies that 
if the variance is upper bounded in the order of $1/\sqrt{T}$, i.e., 
for $k=1,\ldots,K$, 
\begin{equation}
  \frac{1}{T}\sum_{t=0}^{T-1}\left(\Theta_k^{(t)}-\frac{1}{T}\sum_{t'=0}^{T-1}\Theta_k^{(t')}\right)^2\lesssim\frac{1}{\sqrt{T}}, 
  \label{eq:variance}
\end{equation}
$\mathcal{O}(T^\delta)$ becomes equal order to or smaller order than $\mathcal{O}(1/\sqrt{T})$.
Therefore, in this case, 
the convergence rate of the federated algorithm using the proposed deep unfolding-based weights 
is comparable to that of the conventional algorithms summarized in \cite{Book}.
On the other hand, if \eqref{eq:variance} does not hold, 
$\mathcal{O}(T^\delta)$ becomes a leading order 
and the algorithm using the proposed deep unfolding-based weights converges slower than the conventional algorithms.

\begin{figure}[tb]
  \centering
      \includegraphics[width=\columnwidth]{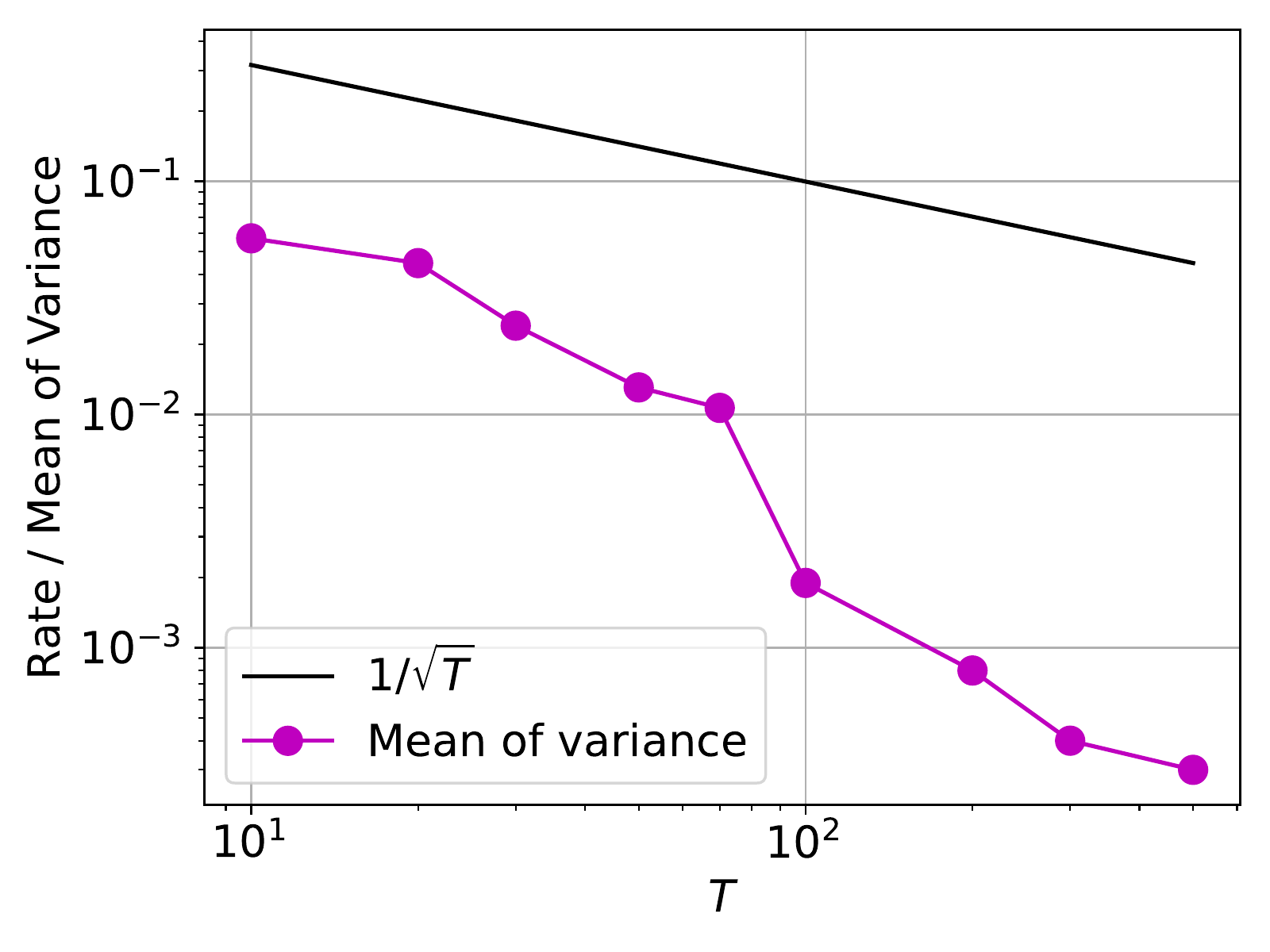}
      \caption{The mean of the variance of learned weights and a line representing $1/\sqrt{T}$.}
      \label{fig:rate}
\end{figure}

We present an example of the variance of learned weights through numerical experiments.
We prepared local datasets for $K=2$ clients extracted from MNIST dataset \cite{mnist} that are IID but unbalanced in quantity.
The experimental codes and datasets are available in a GitHub repository\footnote{https://github.com/a-nakai-k/DeepUnfolding-based-FL}.
One client has $1760$ training data and another has $5739$ data.
The training of deep unfolding-based weights (Algorithm~\ref{alg:prop}) was performed using the datasets, 
and the mean of the variance of the learned weights, i.e., $\sum_{k=1}^2 \sum_{t=0}^{T-1}\left(\Theta_k^{(t)}-\sum_{t'=0}^{T-1}\Theta_k^{(t')}/T\right)^2/T/2$ was calculated for each round $T$.
Figure~\ref{fig:rate} shows the mean of the variance and a line representing $1/\sqrt{T}$.
We can see that the variance was decreasing more rapidly than $1/\sqrt{T}$. 
Therefore, in this case, the order $\mathcal{O}(T^\delta)$ due to the variance of learned weights 
is smaller than $\mathcal{O}(1/\sqrt{T})$, 
and then the leading order of the federated algorithm is $\mathcal{O}(1/\sqrt{T})$.

\section{Experimental Results}

\begin{table*}[tb]
  \caption{Experimental settings for MNIST dataset. Data quantity, number of attached class labels, 
  number of epochs, and communication probability of five clients (clients 0--4).}
  \label{tab:tab}
  \centering
  \begin{tabular}{llllll}
    \hline
    Env. & characteristics & quantity $N_k$ & class label & epochs $E$ & communication\\
    \hline \hline
    $\mathrm{I}$ & quantity skew & $1042, 1023, 862, 1184, 4459$ & all labels & $E=2$ & perfect \\
    $\mathrm{I}\hspace{-1.2pt}\mathrm{I}$ & label distribution skew & $6775, 6774, 6776, 6776, 6776$ & $2,3,5,5,5$ & $E=2$ & perfect  \\
    $\mathrm{I}\hspace{-1.2pt}\mathrm{I}\hspace{-1.2pt}\mathrm{I}$ & computational skew & $1713, 1713, 1713, 1713, 1716$ & all labels & $2,1,1,1,1$ & perfect \\
    $\mathrm{I}\hspace{-1.2pt}\mathrm{V}$ & communication skew & $1713, 1713, 1713, 1713, 1716$ & all labels & $E=2$ & $0.2, 0.3, 0.8, 0.9, 1$ \\
    \hline
  \end{tabular}
 \end{table*}
\subsection{Experimental Setup}
The performance of the proposed approach was evaluated through numerical experiments.
All simulations were executed using Python 3.9.12 and Pytorch 1.12.0.

This paper aims to obtain a model that exhibits high prediction performance on unknown data which do not include class imbalance through the cooperation of clients with non-IID local data.
Therefore, test accuracy was evaluated 
for the model obtained through federated learning 
using class-balanced test data not included in the training data.
The training data of the clients used for the preprocessing and federated learning phases were non-IID and lay in statistically heterogeneous environments.
On the other hand, the test data for performance evaluation were not distributed to the clients.

For the results in the following subsections, 
we assume that 
all clients are cooperative for the proposed preprocessing 
and provide all the data for the preprocessing.

The experimental codes and datasets are available in a GitHub repository\footnotemark[1].

\subsection{Settings for MNIST Dataset}
First, we present the simulation results for small-scale and relatively simple scenarios.
The focus of the simulation is to verify 
that the proposed DUW-FedAvg performs properly 
by determining whether interpretable weights can be obtained.
The number of clients and that of rounds were fixed at $K=5$ and $T=10$, respectively.

\begin{figure*}[tb]
  \begin{tabular}{cccc}
    \begin{minipage}[b]{0.23\hsize}
      \centering
      \includegraphics[width=0.9\columnwidth]{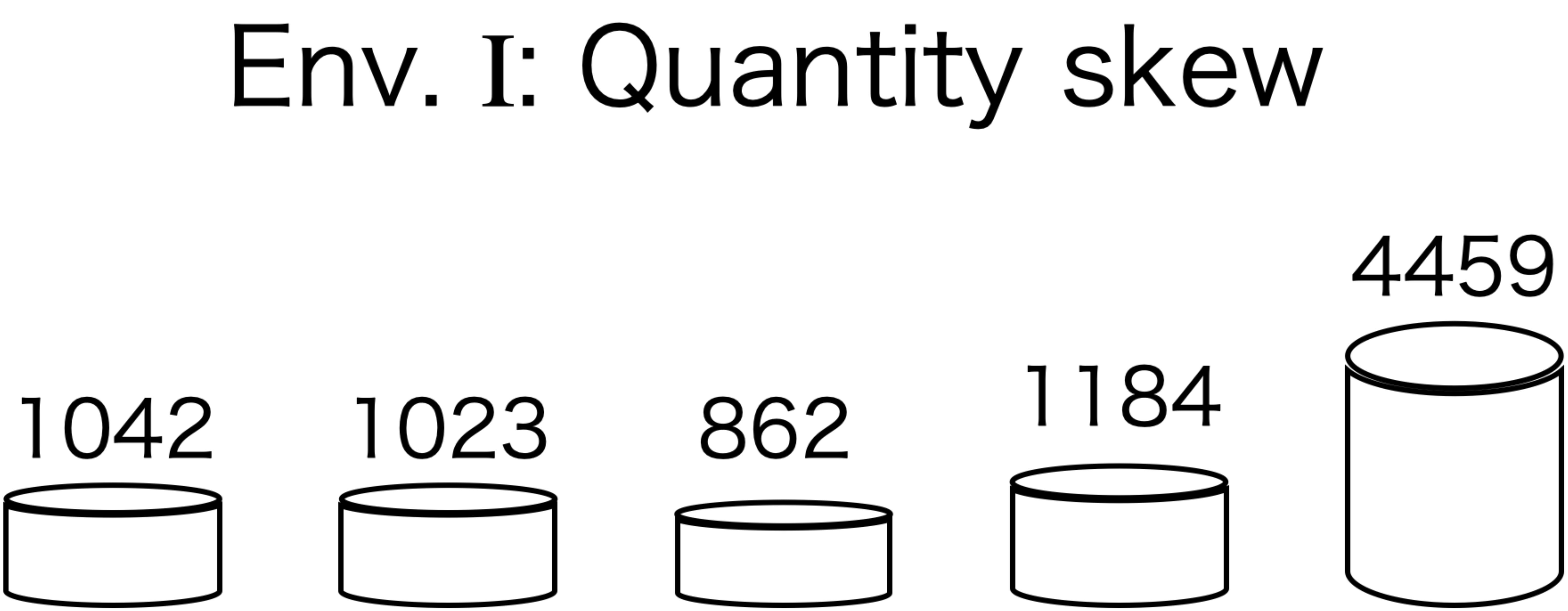}
    \end{minipage} & 
    \begin{minipage}[b]{0.23\hsize}
      \centering
      \includegraphics[width=0.9\columnwidth]{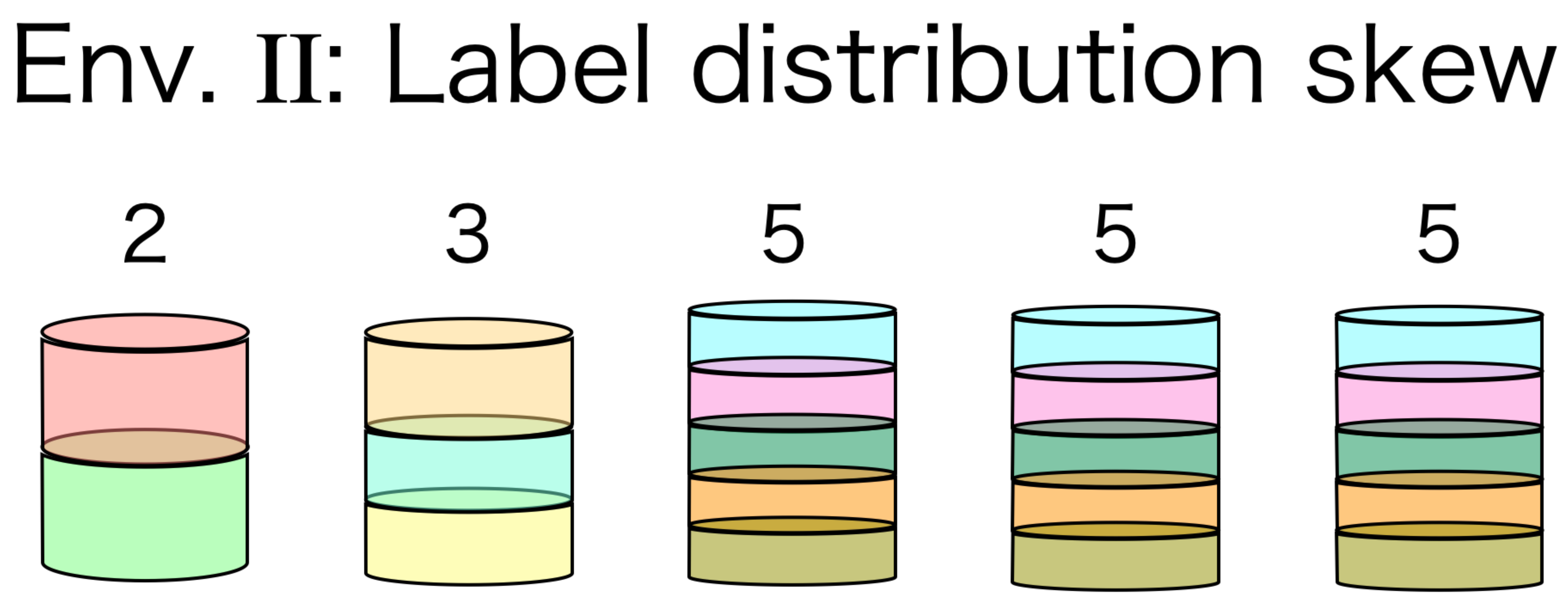}
    \end{minipage} & 
    \begin{minipage}[b]{0.23\hsize}
      \centering
      \includegraphics[width=0.9\columnwidth]{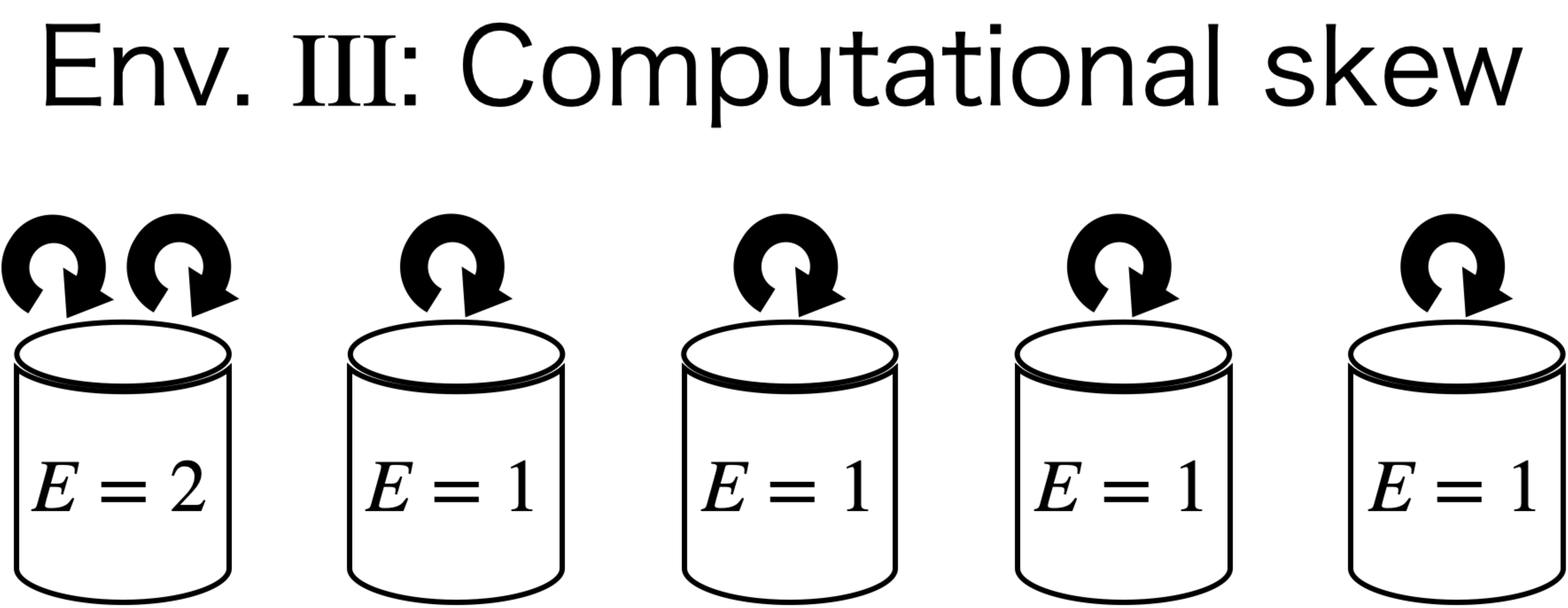}
    \end{minipage} & 
    \begin{minipage}[b]{0.23\hsize}
      \centering
      \includegraphics[width=0.9\columnwidth]{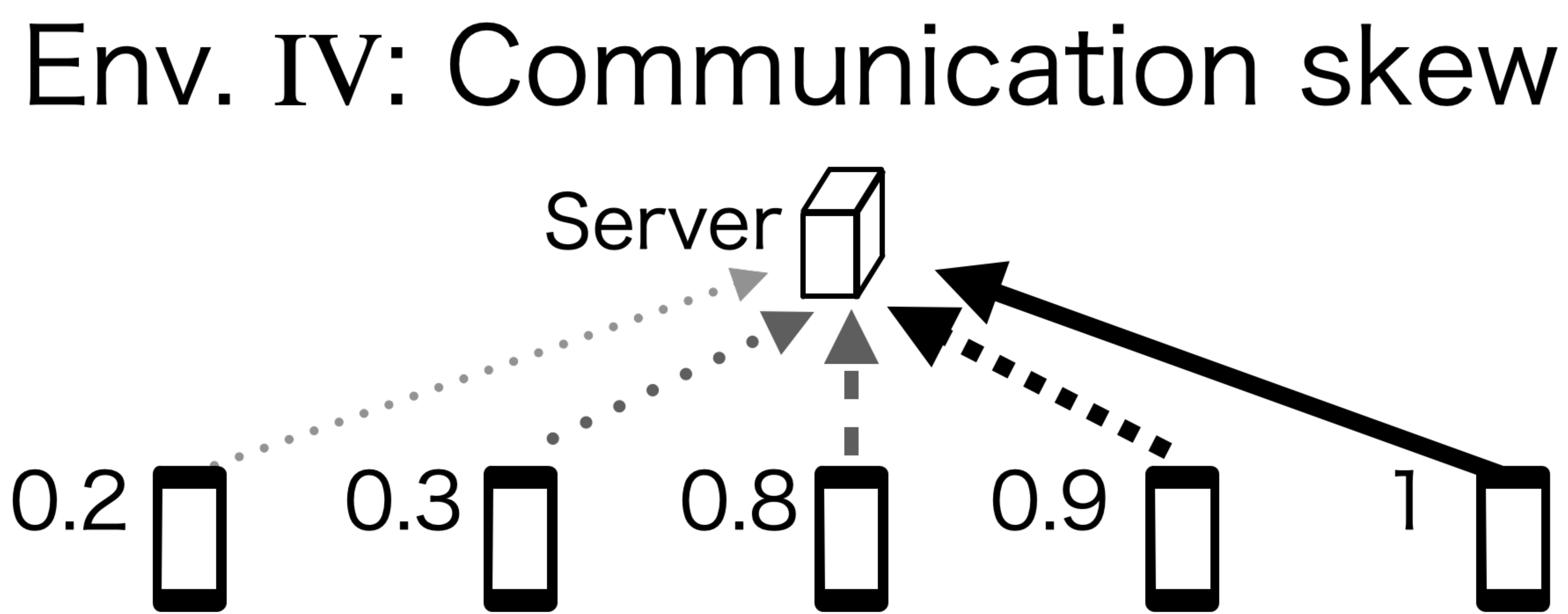}
    \end{minipage}\\
    \begin{minipage}[b]{0.23\hsize}
      \centering
      \includegraphics[width=\columnwidth]{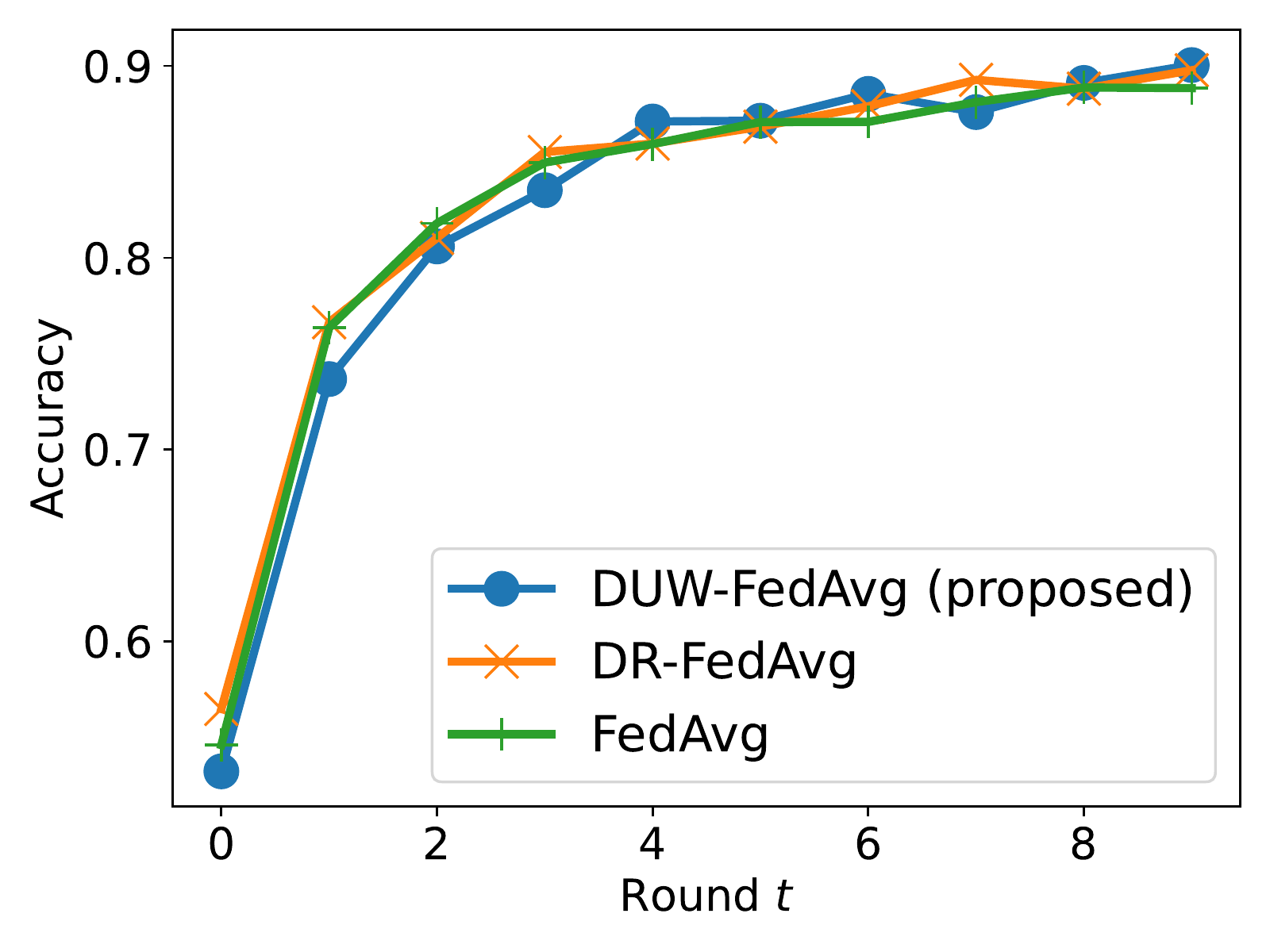}
      \subcaption{Test accuracy (Env.~$\mathrm{I}$)}
    \end{minipage} & 
    \begin{minipage}[b]{0.23\hsize}
      \centering
      \includegraphics[width=\columnwidth]{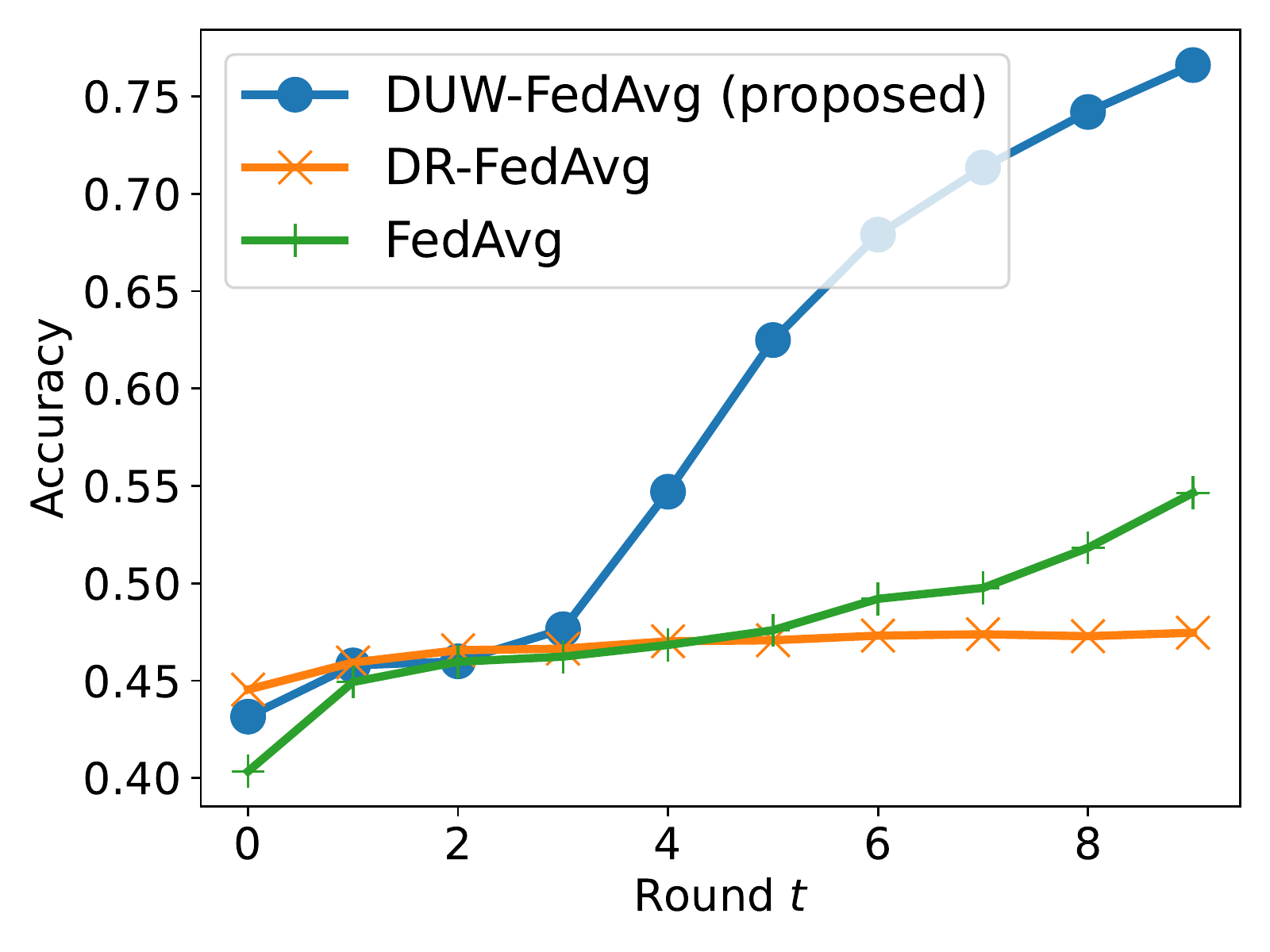}
      \subcaption{Test accuracy (Env.~$\mathrm{I}\hspace{-1.2pt}\mathrm{I}$)}
    \end{minipage} & 
    \begin{minipage}[b]{0.23\hsize}
      \centering
      \includegraphics[width=\columnwidth]{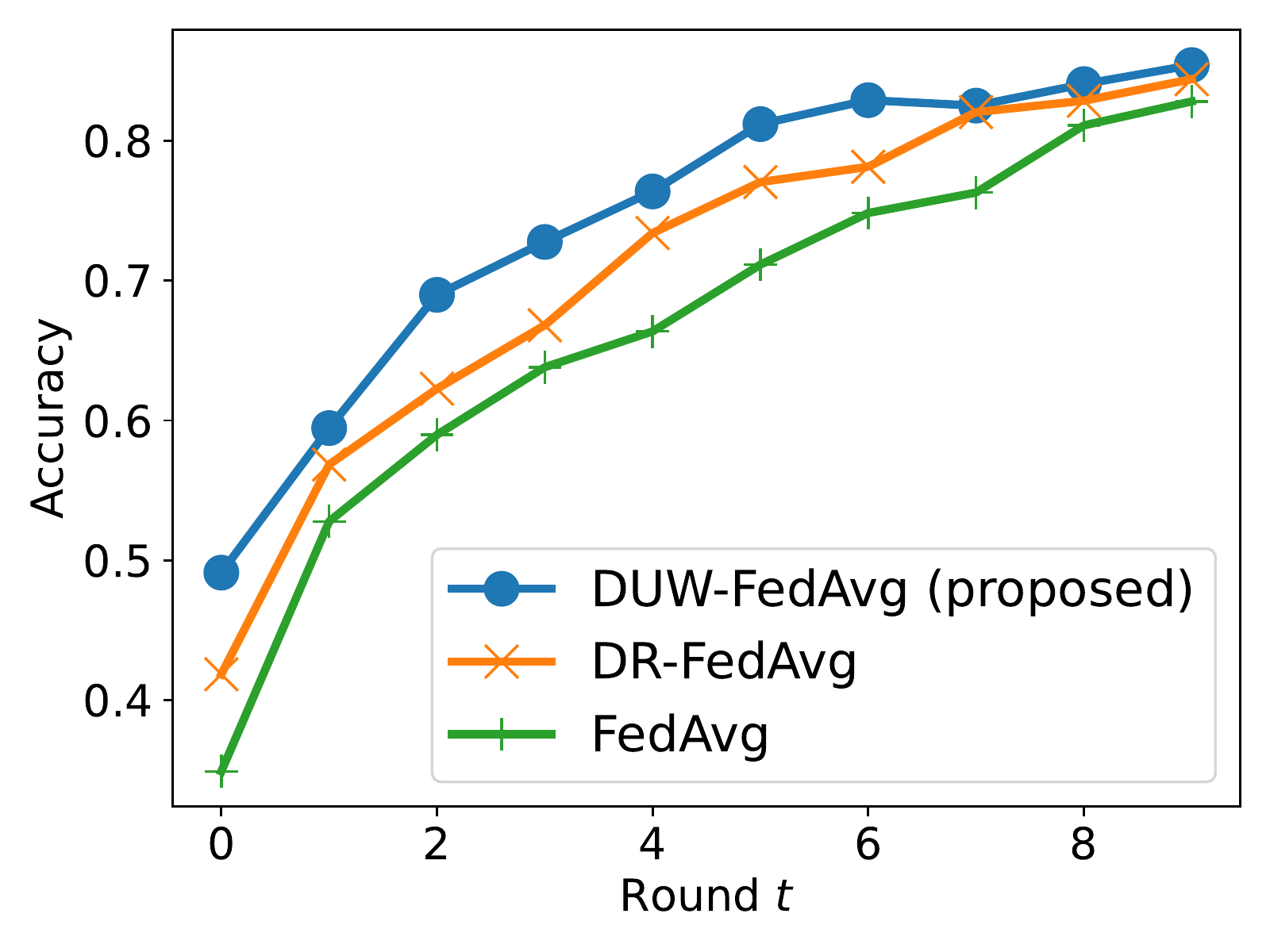}
      \subcaption{Test accuracy (Env.~$\mathrm{I}\hspace{-1.2pt}\mathrm{I}\hspace{-1.2pt}\mathrm{I}$)}
    \end{minipage} & 
    \begin{minipage}[b]{0.23\hsize}
      \centering
      \includegraphics[width=\columnwidth]{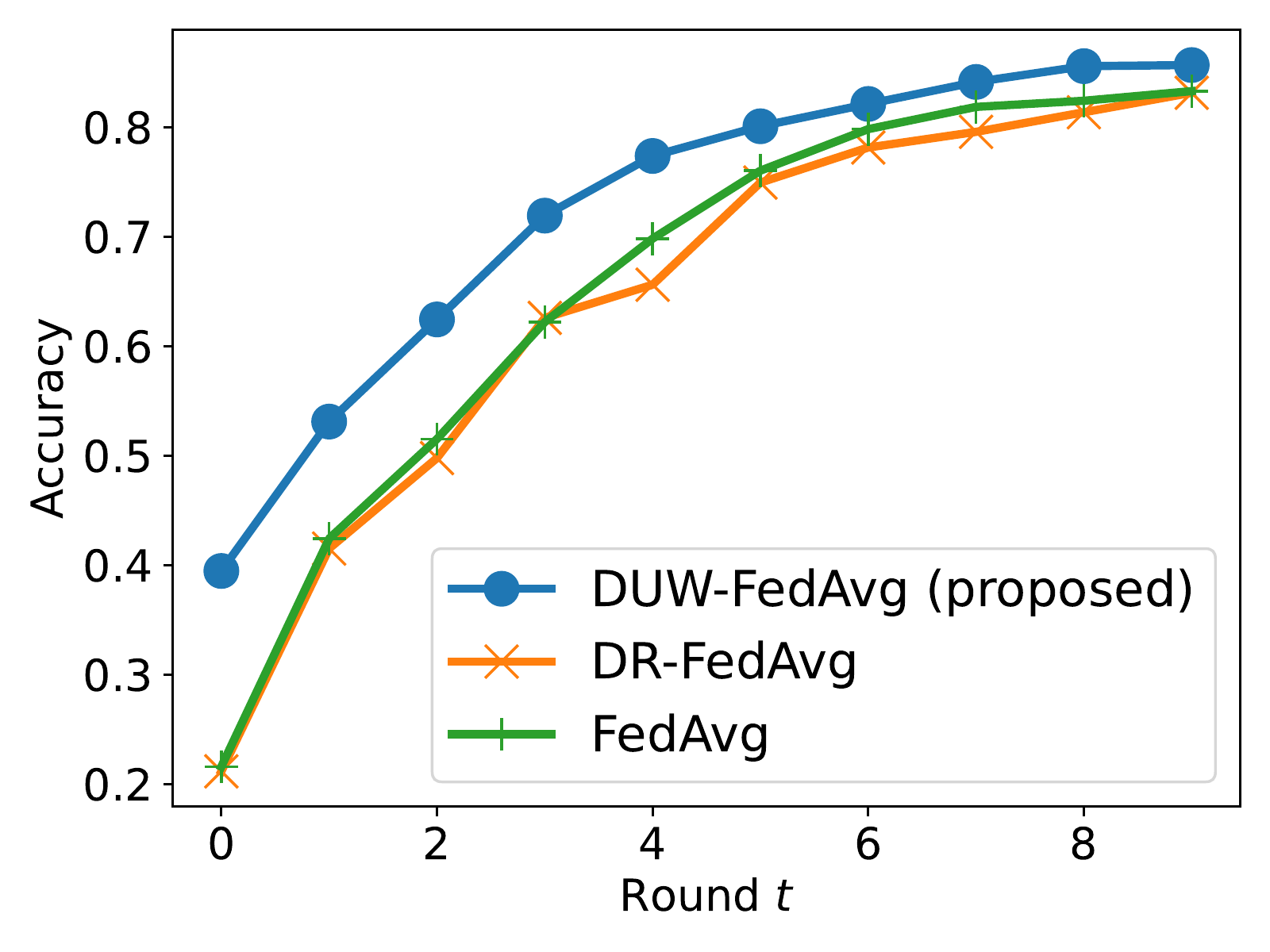}
      \subcaption{Test accuracy (Env.~$\mathrm{I}\hspace{-1.2pt}\mathrm{V}$)}
    \end{minipage}\\
    \begin{minipage}[b]{0.23\hsize}
      \centering
      \includegraphics[width=\columnwidth]{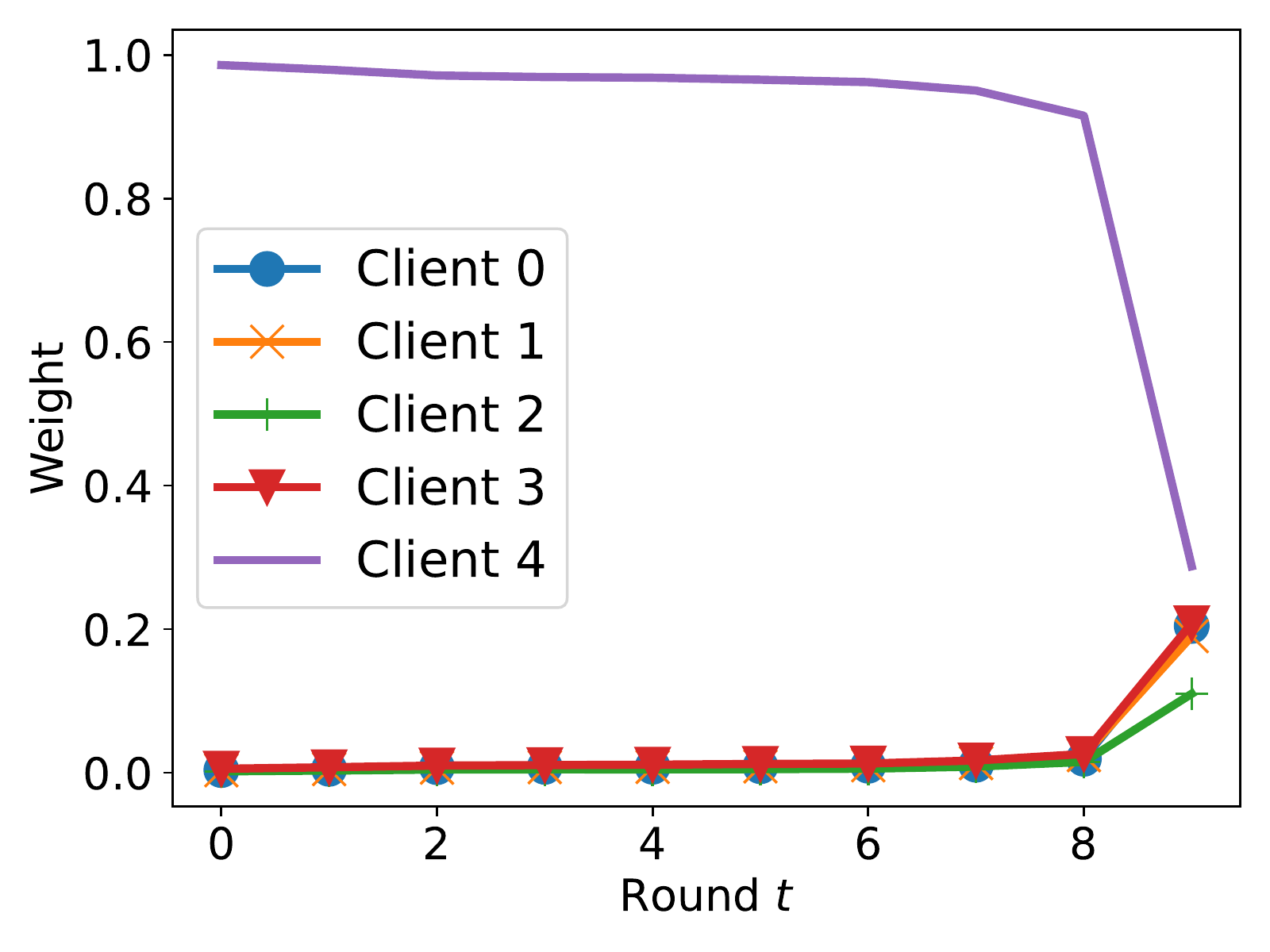}
      \subcaption{Learned weights (Env.~$\mathrm{I}$)}
    \end{minipage} & 
    \begin{minipage}[b]{0.23\hsize}
      \centering
      \includegraphics[width=\columnwidth]{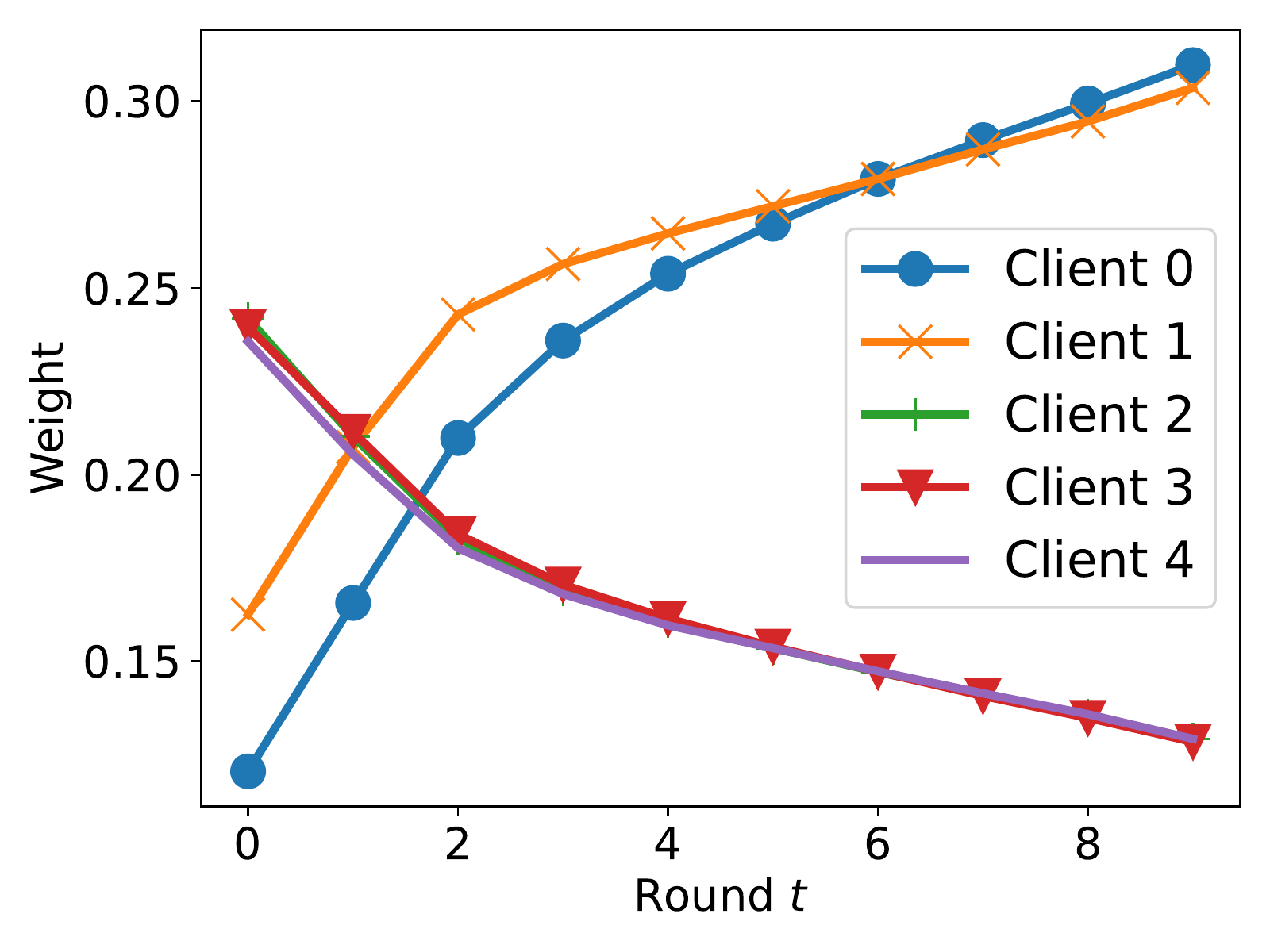}
      \subcaption{Learned weights (Env.~$\mathrm{I}\hspace{-1.2pt}\mathrm{I}$)}
    \end{minipage} & 
    \begin{minipage}[b]{0.23\hsize}
      \centering
      \includegraphics[width=\columnwidth]{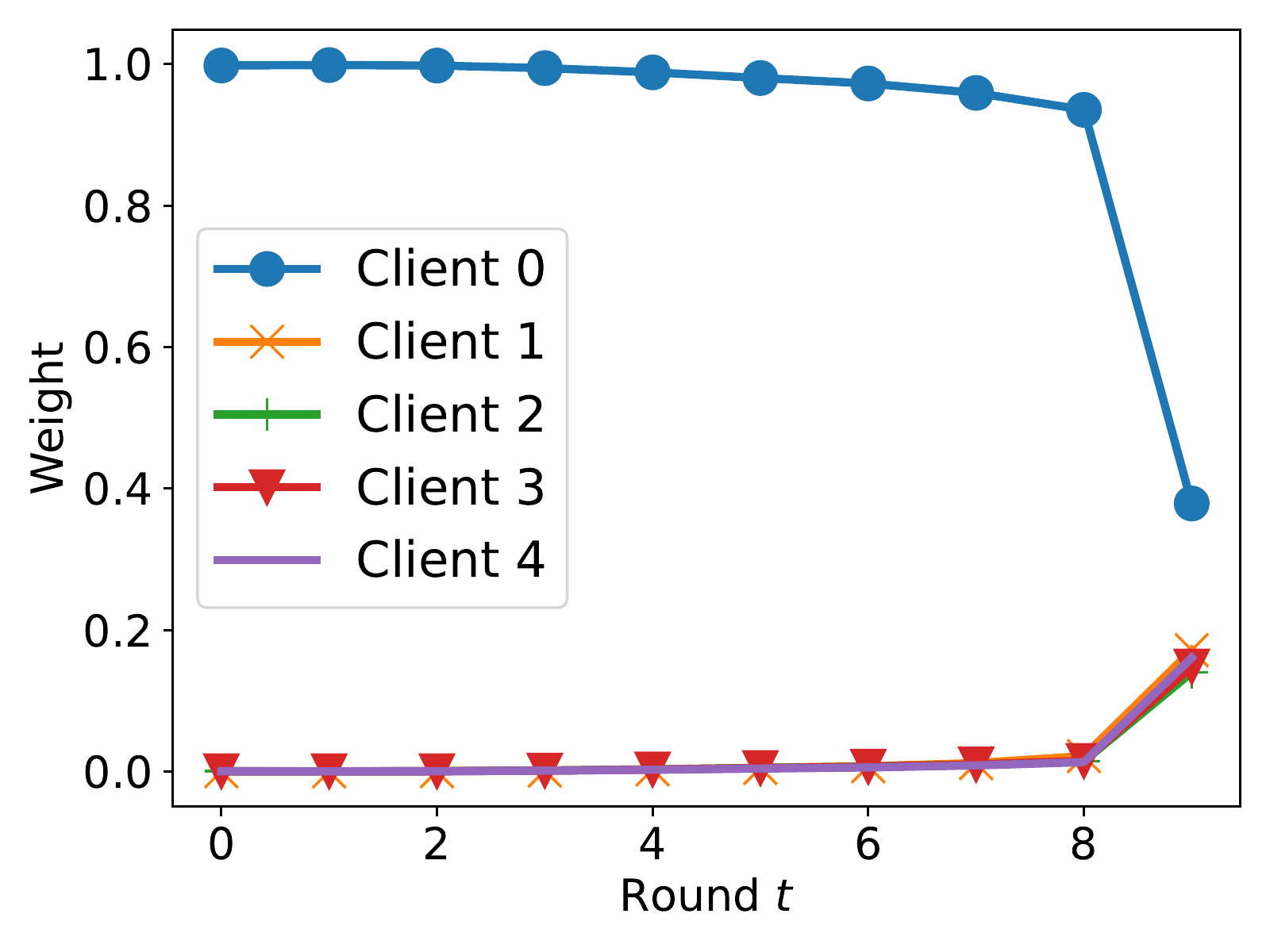}
      \subcaption{Learned weights (Env.~$\mathrm{I}\hspace{-1.2pt}\mathrm{I}\hspace{-1.2pt}\mathrm{I}$)}
    \end{minipage} & 
    \begin{minipage}[b]{0.23\linewidth}
      \centering
      \includegraphics[width=\columnwidth]{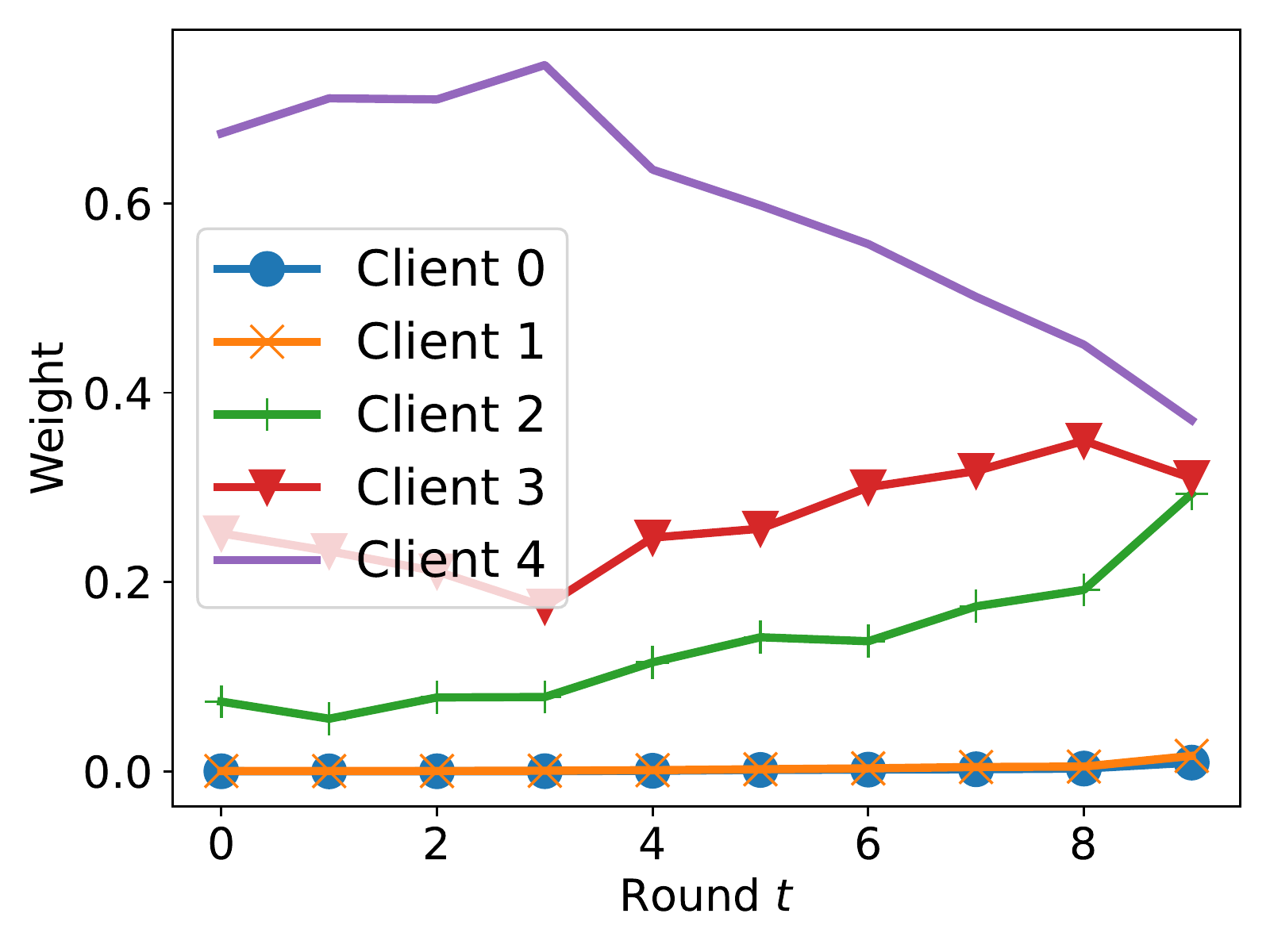}
      \subcaption{Learned weights (Env.~$\mathrm{I}\hspace{-1.2pt}\mathrm{V}$)}
    \end{minipage}
  \end{tabular}
  \caption{Test accuracy and learned weights for environments $\mathrm{I}$ (quantity skew), $\mathrm{I}\hspace{-1.2pt}\mathrm{I}$ (label distribution skew), $\mathrm{I}\hspace{-1.2pt}\mathrm{I}\hspace{-1.2pt}\mathrm{I}$ (computational skew), and $\mathrm{I}\hspace{-1.2pt}\mathrm{V}$ (communication skew).}
  \label{fig:summary}
  \end{figure*}

We prepared four characteristic training datasets that were extracted from the MNIST dataset \cite{mnist} 
(specific parameters are summarized in Table~\ref{tab:tab} and Figure~\ref{fig:summary} top): 
\begin{itemize}
  \item {\bf Environment} $\mathrm{I}$: It contains the quantity skew. 
    Local data are IID but the quantity is not balanced. 
  \item {\bf Environment} $\mathrm{I}\hspace{-1.2pt}\mathrm{I}$: It contains the label distribution skew. 
    The quantity is balanced but each client only has data for specific labels. 
    Specifically, client $0$ has only data of $2$ class labels.
  \item {\bf Environment} $\mathrm{I}\hspace{-1.2pt}\mathrm{I}\hspace{-1.2pt}\mathrm{I}$: It contains the computational capability skew.
    Local data are IID and balanced but the number of epochs $E$ that can calculate during a round varies across clients. 
  \item {\bf Environment}  $\mathrm{I}\hspace{-1.2pt}\mathrm{V}$: It contains the communication capability skew.
    Local data are IID and balanced but each client can transmit the model parameters only with a certain probability.
    For example, client $0$ can transmit its local updates only with probability $0.2$.
\end{itemize}
The model was constructed of three fully connected layers with $128$ features in the hidden layers.
The local minibatch size, learning rate $\mu$ for {\sf ClientUpdate}, and learning rate for the proposed preprocessing phase 
were set to $50, 0.01$, and $0.001$, respectively.
The number of learning iterations was set to $M=400, 400, 500, 1000$ for Envs. $\mathrm{I}$, $\mathrm{I}\hspace{-1.2pt}\mathrm{I}$, $\mathrm{I}\hspace{-1.2pt}\mathrm{I}\hspace{-1.2pt}\mathrm{I}$ and $\mathrm{I}\hspace{-1.2pt}\mathrm{V}$, respectively, 
for sufficient learning.
We used the Adam optimizer and mean squared error (MSE) loss for the proposed method.
We compare the accuracy for the test dataset (class-balanced $10000$ data, not included in the training datasets) 
of the original FedAvg \cite{McMahan} with \eqref{eq:thetafedavg}, DR-FedAvg \cite{Zhao} with \eqref{eq:thetadr}, and the proposed DUW-FedAvg with the learned weights.
The initial weights of the proposed training process were set as in \eqref{eq:thetafedavg}.
The hyperparameter for DR-FedAvg was set to $q=1$.

\subsection{Results on MNIST Dataset}
The test accuracy and learned weights obtained using the proposed method are shown in Figure~\ref{fig:summary}.

Figure~\ref{fig:summary} (a) shows the accuracy for Env. $\mathrm{I}$.
A few performance differences were observed between the methods. 
This is because the original FedAvg can address the quantity skew, as observed in \eqref{eq:objective}.
The learned weights corresponding to each round and each client are shown in Figure~\ref{fig:summary} (e).
The figure indicates that a larger value was assigned to client $4$, which has more data than the others.
This behavior coincided with the weights of FedAvg.
The information of client $4$ was emphasized in most rounds, 
but, in the final round, the model parameters of all clients were almost equally aggregated.

Figure~\ref{fig:summary} (b) shows the accuracy for Env. $\mathrm{I}\hspace{-1.2pt}\mathrm{I}$.
The accuracy of DR-FedAvg and FedAvg was only $\leq55$\%.
However, the proposed DUW-FedAvg achieved an accuracy of $75$\%. 
In other words, the proposed method is suitable for adapting to the label distribution skew 
that cannot be addressed using conventional methods.
The learned weights shown in Figure~\ref{fig:summary} (f) indicate that 
all clients' information was equally aggregated in the first few rounds 
but, as the federate learning process progressed, the information of clients with unique labels was emphasized.

Figure~\ref{fig:summary} (c) shows the accuracy for Env. $\mathrm{I}\hspace{-1.2pt}\mathrm{I}\hspace{-1.2pt}\mathrm{I}$.
The proposed DUW-FedAvg achieves higher accuracy in each round than the other methods. 
The behavior of the learned weights shown in Figure~\ref{fig:summary} (g) is similar to that shown in Figure~\ref{fig:summary} (e), 
where a larger value was assigned to client $0$, which had a higher computational capability than the others.

Figure~\ref{fig:summary} (d) shows the accuracy for Env. $\mathrm{I}\hspace{-1.2pt}\mathrm{V}$.
The proposed DUW-FedAvg achieved a higher accuracy in each round than the other methods. 
The learned weights shown in Figure~\ref{fig:summary} (h) indicate that, 
primarily in the first half of the rounds, weights were assigned depending on the communication probability of the clients.
We can see that the proposed method can also adapt to the communication capability of the clients.

In summary, a larger weight should be assigned to client 
with larger amount of data, data with more unique labels, higher computational capability, and higher communication capability.
These results are consistent with general intuition. 
Thus, we can see that the proposed method yields interpretable behavior.

\begin{figure}[tb]
  \centering
      \includegraphics[width=\columnwidth]{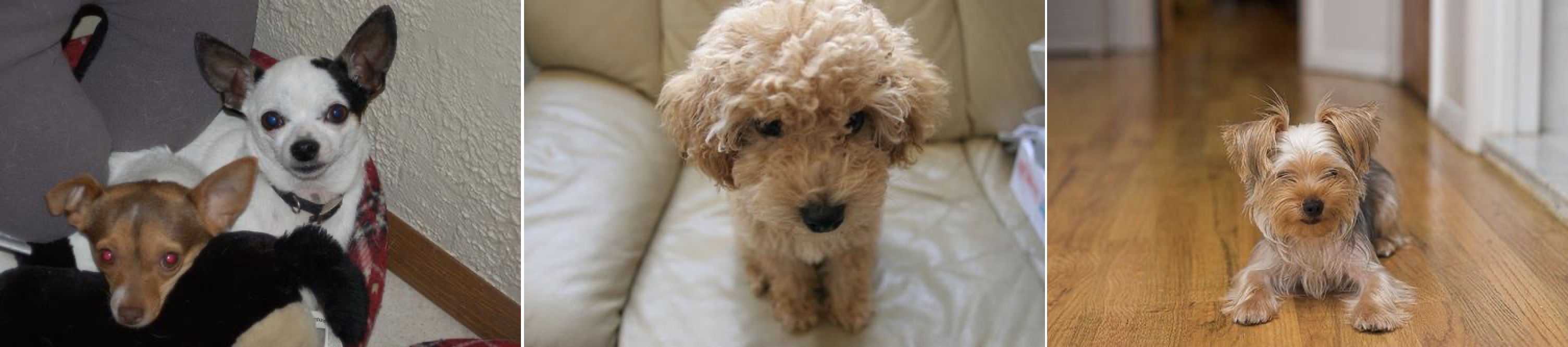}
      \caption{Examples of Stanford Dogs dataset \cite{dogs1,dogs2}: Chihuahua, toy poodle, and Yorkshire terrier.}
      \label{fig:dogs}
\end{figure}
\subsection{Settings for Stanford Dogs Dataset}
\begin{table*}[thb]
  \caption{Experimental settings for Stanford Dogs dataset.}
  \label{tab:tab2}
  \centering
  \centering
        \begin{tabular}{l|l|lllll}\hline
        \multicolumn{2}{l|}{Client} & 0 & 1 & 2 & 3 & 4 \\ \hline
            Data property & \#train data & 3295 & 3345 & 1868 & 3178 & 4773 \\
            & \#class labels & 24 & 25 & 18 & 25 & 39 \\ \hline
            Environment A & communication probability & 0.9600 & 0.6995 & 0.9999 & 0.2201 & 0.3611\\ 
            & \#epochs & 5 & 5 & 5 & 5 & 5\\ \hline
            Environment B & communication probability & 0.5173 & 0.9470 & 0.7655 & 0.2824 & 0.2210\\ 
            & \#epochs & 4 & 7 & 1 & 2 & 7\\ \hline
        \end{tabular}
\end{table*}
We also evaluated the performance of the proposed method using another dataset with a more complex heterogeneity: 
image classification using the Stanford Dogs dataset \cite{dogs1,dogs2}.
The dataset contains images of $120$ dog breeds.
Figure~\ref{fig:dogs} shows examples of the dataset.

\begin{figure}[tb]
  \centering
      \includegraphics[width=\columnwidth]{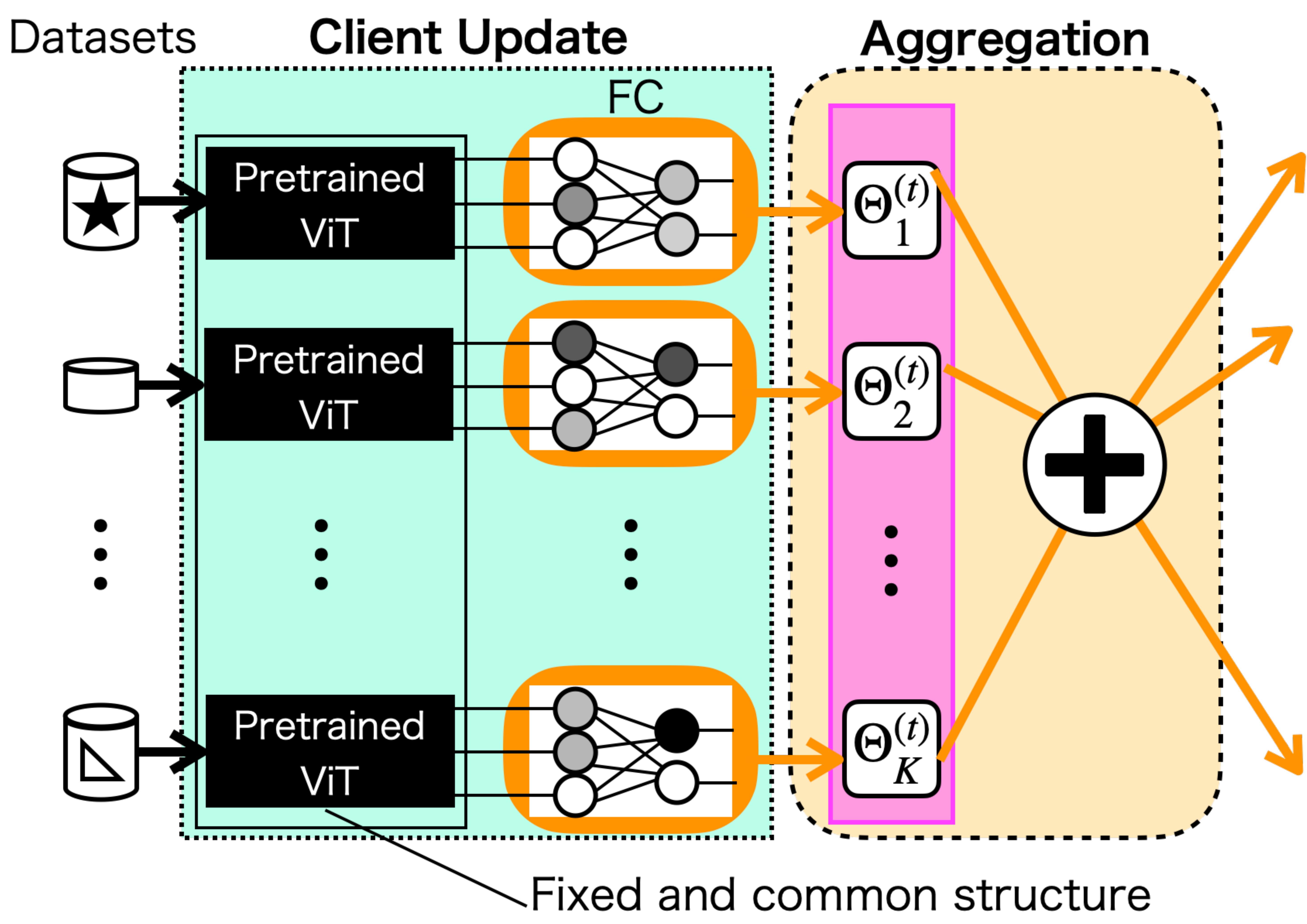}
      \caption{Model for classification of Stanford Dogs dataset with pretrained ViT $+$ fully connected (FC) layer.}
      \label{fig:vit}
\end{figure}
For image classification using the real dataset, 
we adopted a model composed of ViT and one fully connected layer, as shown in Figure~\ref{fig:vit}, 
and we used transfer learning.
We employed a pretrained ViT model on ImageNet-21k (14 million images, 21,843 classes, at a resolution of 224$\times$224 pixels) \cite{vit}.
The ViT model is shared by all the clients.
The number of model parameters of ViT is 86 million, the value of which is fixed for this experiment. 
Only the model parameters of the fully connected layer are updated.
Such a handling of large-scale models in federated learning using pretrained models is discussed in \cite{Tanpretrained} 
and the pretraining using federated learning is proposed in \cite{Tian}.

We verified the performance of the proposed approach in complex combinations of statistical and device heterogeneity.
The specific properties of data and device heterogeneity settings are summarized in Table~\ref{tab:tab2}.
We assumed $K=5$ clients and constructed non-IID datasets using the Dirichlet distribution \cite{mnist,Hsu,Yurochkin,Lin}.
The concentration parameter of the Dirichlet distribution was set to $0.0005$, 
where a smaller value promotes clients holding data for specific classes and enhances non-IIDness.
The number of tranining data was $16459$ and that of the test data was $4116$.
In addition, we established the following two environments for device heterogeneity:
\begin{itemize}
  \item {\bf Environment} A: It contains device heterogeneity.
    Each client can transmit the model parameters only with a certain probability.
    The communication probabilities were generated randomly.
  \item {\bf Environment} B: It contains device heterogeneity.
    The number of epochs $E$ that can calculate during a round varies across clients. 
    Each client can transmit the model parameters only with a certain probability. 
    The number of epochs that each client can perform in a round 
    and the communication probabilities were generated randomly.
\end{itemize}

The local minibatch size, learning rate $\mu$ for {\sf ClientUpdate} 
and {\sf ClientUpdateNova}, 
number $M$ of learning iterations, 
and learning rate for the proposed preprocessing phase 
were set to $100, 0.01$, $1000$, and $0.001$, respectively.
The number $T$ of rounds was set to $T=100$ for Environment A 
and $T=70$ for Environment B.
We used the Adam optimizer and MSE loss for the proposed method.

We compare the accuracy for the test dataset of the following seven methods.
The original FedAvg \cite{MATCHA} and FedNova \cite{FedNova} are the conventional federated learning algorithms.
DR-FedAvg \cite{Zhao}, FedAdp \cite{FedAdp}, and FedFa \cite{FedFa} are the conventional weighting strategies.
DUW-FedAvg and DUW-FedNova are the proposed methods which are combined with the original FedAvg and FedNova, respectively.
The conventional FedFa includes local updates with momentum, 
but we used vanilla SGD instead of this to make it comparable to the other methods.
The hyperparameters for DR-FedAvg, FedAdp, and FedFa were $q=1$, $\beta=7$, and $\gamma=0.5$, respectively.
The initial weights of the proposed training process were set as in \eqref{eq:thetafedavg}.

The proposed preprocessing phase, i.e., the training of the deep unfolding-based weight, 
was performed only once for each experimental environment.
Subsequently, the federated learning phase and test phase were performed $10$ times in each environment. 
We then calculate the average test accuracy over the $10$ trials.

\subsection{Results for Stanford Dogs Dataset}
\begin{table}[tb]
  \caption{Average test accuracy (\%) on Stanford Dogs dataset.}
  \label{tab:tab3}
  \centering
  \begin{tabular}{l|ll}
    \hline
    \multicolumn{1}{c|}{} & \multicolumn{2}{c}{Environment}\\
    \cline{2-3}
    Method & A & B\\
    \hline \hline
    Original FedAvg \cite{McMahan} & $81.63 \pm 2.24$ & $58.32\pm 2.62$ \\
    FedNova \cite{FedNova} & $76.93\pm 2.89$ & $\mathbf{71.00\pm5.34}$ \\
    \hline
    DR-FedAvg \cite{Zhao} & $87.22\pm 0.30$ & $61.37\pm 1.22$  \\
    FedAdp \cite{FedAdp} & $78.67\pm 2.26$ & $61.45\pm11.61$ \\
    FedFa \cite{FedFa} & $77.92\pm 2.58$ & $60.56\pm15.24$ \\
    \hline\hline
    Proposed DUW-FedAvg & $\mathbf{88.15\pm 0.63}$ & $62.51\pm 0.48$ \\
    Proposed DUW-FedNova & $85.73\pm 1.97$ & $60.34\pm1.68$ \\
    \hline
  \end{tabular}
\end{table}
Table~\ref{tab:tab3} lists the average test accuracy and standard deviation obtained using the conventional and proposed methods.

With respect to the proposed DUW-FedAvg, 
the proposed approach improved the performance of the original FedAvg in both Environments A and B, 
and achieved better performance than the conventional weighting strategies, DR-FedAvg, FedAdp, and FedFa.
DUW-FedAvg achieved the highest accuracy among all the methods in Environment A. 
The conventional FedAdp and FedFa caused unstable test performance in Environment B, 
which was obvious from the large standard deviations.
On the other hand, 
the proposed DUW-FedAvg provided relatively stable performance.

In case of the proposed DUW-FedNova, 
the performance was improved over FedNova by the proposed approach in Environment A.
However, in Environment B, 
the performance has deteriorated.
It can be understood 
that application of the proposed approach to FedNova is somewhat more difficult than that to FedAvg, 
and there is room for improvement in the setting of loss function or training procedure.

We can conclude that 
the proposed approach yields performance improvement, especially for FedAvg, even in image classification task with a large-scale model.
It means that 
the proposed approach can be applicable to practical real-world tasks.

\section{Conclusion}
This paper proposed deep unfolding-based weighted averaging for federated learning 
that can directly incorporate heterogeneity in an environment of interest.
The proposed approach can be combined with various federated learning algorithms 
and can flexibly adapt to the complex heterogeneity of the environment.
We first evaluated the performance of the proposed DUW-FedAvg 
under the assumptions of specific environments and reliable clients.
The proposed algorithm achieved better accuracy than the conventional methods, 
particularly in environments with label distribution skew and computational/communication capability skew.
The learned weights behaved as adapting to the heterogeneity.
The proposed methods also could be applied to image classification task using a large-scale model.
This implies that the proposed approach has sufficient potential for handling practical tasks in real worlds.
We have theoretically shown that 
the convergence rate of federated learning algorithm with the proposed method becomes comparable to that of the conventional algorithms 
as long as the variance of learned weights is bounded.

The proposed approach requires the overhead of the pre-training process, 
and the computational load can be severe owing to practical reasons such as memory constraints.
Therefore, improvements in computational efficiency are required 
to apply this approach to a large-scale federated learning. 
For example, the cross-device context of federated learning 
comprising many unreliable clients must be considered. 
The improvement of the method's scalability and robustness by combining it with client selection 
will be addressed in a future study.
We can also expect that 
interpretable behavior of the learned weights shown in this paper 
leads to the derivation of a novel adaptive weighting strategy 
with scalability and robustness.

\section*{Appendix}
We prove Theorem~\ref{theo:convergence}.
For convenience, we define the following gradient: 
\begin{equation}
  \bm{h}_k^{(t)} = \frac{1}{\|\bm{a}_k^{(t)}\|_1}\sum_{\tau=0}^{\tau_k^{(t)}-1}a_{k,\tau}^{(t)}\nabla f_k(\bm{w}_{k,t}^\tau), 
\end{equation}
where $\bm{a}_k^{(t)}=[a_{k,0}^{(t)},\ldots,a_{k,\tau_k^{(t)}-1}^{(t)}]^\mathrm{T}$ and $a_{k,\tau}^{(t)}\geq0$.
By considering the definition of $\bm{d}_k^{(t)}$ \eqref{eq:dktnew}, 
$\mathbb{E}[\bm{d}_k^{(t)}-\bm{h}_k^{(t)}]=0$ and $\mathbb{E}[\langle\bm{d}_k^{(t)}-\bm{h}_k^{(t)},\bm{d}_l^{(t)}-\bm{h}_l^{(t)}\rangle]=0 \ (k\neq l)$ hold.
From the assumption of Lipschitz-smootheness, 
\begin{align}
  &\mathbb{E}\left[f(\bm{w}^{(t+1)})\right]-f(\bm{w}^{(t)}) \nonumber \\
  &\leq-\eta\tau_{\mathrm{eff}}^{(t)}\mathbb{E}\left[\langle\nabla f(\bm{w}^{(t)}),\sum_k \Theta_k^{(t)}\bm{d}_k^{(t)}\rangle\right] \nonumber \\
  &\quad+\frac{\eta^2(\tau_{\mathrm{eff}}^{(t)})^2L}{2}\mathbb{E}\left[\left\|\sum_k\Theta_k^{(t)}\bm{d}_k^{(t)}\right\|^2\right].
\end{align}
This expectation operates in terms of minibatches.
From (62) of \cite{FedNova}, we can expand the bound as follows:
\begin{align}
  &\mathbb{E}\left[f(\bm{w}^{(t+1)})\right]-f(\bm{w}^{(t)}) \nonumber \\
  &\leq\frac{-\tau_{\mathrm{eff}}^{(t)}\eta}{2}\!\left[\left\|\nabla f(\bm{w}^{(t)})\right\|^2\right] \!-\! \frac{\tau_{\mathrm{eff}}^{(t)}\eta(1-\tau_{\mathrm{eff}}^{(t)}\eta L)}{2}\!\left[\left\|\Theta_k^{(t)}\bm{h}_k^{(t)}\right\|^2\right] \nonumber \\
  &\quad+(\tau_{\mathrm{eff}}^{(t)})^2\eta^2L\sigma^2\sum_k\frac{(\Theta_k^{(t)})^2\|\bm{a}_k^{(t)}\|^2}{\|\bm{a}_k^{(t)}\|_1^2} \nonumber \\
  &\quad+\frac{\tau_{\mathrm{eff}}^{(t)}\eta}{2}\mathbb{E}\left[\left\|\nabla f(\bm{w}^{(t)})-\sum_k\Theta_k^{(t)}\bm{h}_k^{(t)}\right\|^2\right].
\end{align}
When $1-\tau_{\mathrm{eff}}^{(t)}\eta L\geq0$, the second term on the right-hand side becomes nonpositive and negligible.
We then obtain 
\begin{align}
  &\frac{\mathbb{E}\left[f(\bm{w}^{(t+1)})\right]-f(\bm{w}^{(t)})}{\tau_{\mathrm{eff}}^{(t)}\eta} \nonumber \\
  &\leq-\frac{1}{2}\left\|\nabla f(\bm{w}^{(t)})\right\|^2+\tau_{\mathrm{eff}}^{(t)}\eta L\sigma^2\sum_k\frac{(\Theta_k^{(t)})^2\|\bm{a}_k^{(t)}\|^2}{\|\bm{a}_k^{(t)}\|_1^2} \nonumber \\
  &\quad+\underset{\mathcal{T}_1}{\underline{\frac{1}{2}\mathbb{E}\left[\left\|\nabla f(\bm{w}^{(t)})-\sum_k\Theta_k^{(t)}\bm{h}_k^{(t)}\right\|^2\right]}}.
  \label{eq:tmp}
\end{align}
Through the relation $\|\bm{a}+\bm{b}\|^2\leq2\|\bm{a}\|^2+2\|\bm{b}\|^2$, 
\begin{align}
  \mathcal{T}_1\leq&\mathbb{E}\left[\left\|\sum_k\theta_k(\nabla f_k(\bm{w}^{(t)})-\bm{h}_k^{(t)})\right\|^2\right] \nonumber \\
  &+\underset{\mathcal{T}_2}{\underline{\mathbb{E}\left[\left\|\sum_k(\Theta_k^{(t)}-\theta_k)\bm{h}_k^{(t)}\right\|^2\right]}}.
  \label{eq:tmp2}
\end{align}

We can expand the bound in terms of the first term on the right-hand side using relation (87) in \cite{FedNova}: 
\begin{align}
  &\mathbb{E}\left[\left\|\sum_k\theta_k(\nabla f_k(\bm{w}^{(t)})-\bm{h}_k^{(t)})\right\|^2\right] \nonumber \\
  &\leq \frac{2\eta^2L^2\sigma^2}{1-D}\sum_{k=1}^K \theta_k(\|\bm{a}_k^{(t)}\|^2-(a_{k,\tau_k^{(t)}-1}^{(t)})^2) \nonumber \\
  &\quad+ \frac{D\beta^2}{1-D}\left\|\nabla f(\bm{w}^{(t)})\right\|^2 + \frac{D\kappa^2}{1-D}.
  \label{eq:tmp3}
\end{align}
where $D=4\eta^2L^2\max_k\{\|\bm{a}_k^{(t)}\|_1(\|\bm{a}_k^{(t)}\|_1-a_{k,\tau_k^{(t)}-1}^{(t)})\}$. 
By substituting \eqref{eq:tmp2} and \eqref{eq:tmp3} into \eqref{eq:tmp} and rearranging them, we obtain  
\begin{align}
  &\frac{\mathbb{E}\left[f(\bm{w}^{(t+1)})\right]-f(\bm{w}^{(t)})}{\tau_{\mathrm{eff}}^{(t)}\eta} \nonumber \\
  &\leq-\frac{1-D(1+2\beta^2)}{2(1-D)}\left\|\nabla f(\bm{w}^{(t)})\right\|^2 \nonumber \\
  &\quad+ \tau_{\mathrm{eff}}^{(t)}\eta L\sigma^2\sum_k\frac{(\Theta_k^{(t)})^2\|\bm{a}_k^{(t)}\|^2}{\|\bm{a}_k^{(t)}\|_1^2} \nonumber \\
  &\quad+\frac{2\eta^2L^2\sigma^2}{1-D}\sum_k\theta_k(\|\bm{a}_k^{(t)}\|^2-(a_{k,\tau_k^{(t)}-1}^{(t)})^2)+\frac{D\kappa^2}{1-D}+\mathcal{T}_2.
\end{align}

When $D\leq1/(4\beta^2+1)$ holds, then $1/(1-D)\leq1+1/4\beta^2\leq5/4$ and $2D\beta^2/(1-D)\leq1/2$ also hold.
Using these inequalities, we obtain 
\begin{align}
  &\frac{\mathbb{E}\left[f(\bm{w}^{(t+1)})\right]-f(\bm{w}^{(t)})}{\tau_{\mathrm{eff}}^{(t)}\eta} \nonumber \\
  &\leq-\frac{1}{4}\left\|\nabla f(\bm{w}^{(t)})\right\|^2 +\tau_{\mathrm{eff}}^{(t)}\eta L\sigma^2\sum_k\frac{(\Theta_k^{(t)})^2\|\bm{a}_k^{(t)}\|^2}{\|\bm{a}_k^{(t)}\|_1^2} \nonumber \\
  &\quad+\frac{5\eta^2L^2\sigma^2}{2}\sum_k\theta_k(\|\bm{a}_k^{(t)}\|^2-(a_{k,\tau_k^{(t)}-1}^{(t)})^2) \nonumber \\
  &\quad+5\eta^2L^2\kappa^2\max_k\{\|\bm{a}_k^{(t)}\|_1(\|\bm{a}_k^{(t)}\|_1-a_{k,\tau_k^{(t)}-1}^{(t)})\}+\mathcal{T}_2.
\end{align}
Taking the average across all rounds, we obtain 
\begin{align}
  &\frac{1}{T}\sum_t\mathbb{E}\left[\left\|\nabla f(\bm{w}^{(t)})\right\|^2\right] \nonumber \\
  &\leq\frac{4}{T\tau_{\mathrm{eff}}\eta}(f(\bm{w}^{(0)})-f_\mathrm{inf})+\frac{4\eta L\sigma^2\tilde{A}}{K} \nonumber \\
  &\quad+10\eta^2L^2\sigma^2\tilde{B}+20\eta^2L^2\kappa^2\tilde{C}+4\cdot\underset{\mathcal{T}_3}{\underline{\frac{1}{T}\sum_t\mathbb{E}[\mathcal{T}_2]}}, 
  \label{eq:tmp6}
\end{align}
where the coefficients $\tilde{A},\tilde{B}$ and $\tilde{C}$ are defined as 
\begin{equation}
  \tilde{A} = \frac{1}{T}\sum_{t=0}^{T-1}K\sum_{k=1}^K\frac{\left(\Theta_k^{(t)}\right)^2\|\bm{a}_k^{(t)}\|^2}{\|\bm{a}_k^{(t)}\|_1^2},
  \label{eq:Atilde}
\end{equation}
\begin{equation}
  \tilde{B} = \frac{1}{T}\sum_{t=0}^{T-1}\theta_k(\|\bm{a}_k^{(t)}\|^2-(a_{k,-1}^{(t)})^2), 
  \label{eq:Btilde}
\end{equation}
\begin{equation}
  \tilde{C} = \frac{1}{T}\sum_{t=0}^{T-1}\max_k\left(\|\bm{a}_k^{(t)}\|_1(\|\bm{a}_k^{(t)}\|_1-a_{k,\tau_k^{(t)}-1}^{(t)})\right),
  \label{eq:Ctilde}
\end{equation}
$\tau_{\mathrm{eff}}=\sum_t \tau_{\mathrm{eff}}^{(t)}/T$, 
and $f_{\mathrm{inf}}$ is the objective function value after iterating sufficient rounds.

In terms of the last term $\mathcal{T}_3$ of the right-hand side of \eqref{eq:tmp6}, 
\begin{align}
  \mathcal{T}_3&\leq\frac{2^{K-1}}{T}\sum_{k=1}^K\sum_{t=0}^{T-1}(\Theta_k^{(t)}-\theta_k)^2\mathbb{E}\left[\|\bm{h}_k^{(t)}\|^2\right],
  \label{eq:tmp4}
\end{align}
where the inequality follows from $\|\sum_{k=1}^K \bm{v}_k\|^2\leq2^{K-1}\sum_{k=1}^K\|\bm{v}_k\|^2$ for the vectors $\{\bm{v}_k\}$.
From the definition of $\bm{h}_k^{(t)}$ and Assumptions \ref{ass:sigma} and \ref{ass:G}, 
\begin{align}
  &\mathbb{E}\left[\|\bm{h}_k^{(t)}\|^2\right] \nonumber \\
  &\leq2\mathbb{E}\left[\|\bm{h}_k^{(t)}-\bm{d}_k^{(t)}\|^2\right]+2\mathbb{E}\left[\|\bm{d}_k^{(t)}\|^2\right] \nonumber \\
  &=2\mathbb{E}\left[\left\|\frac{1}{\|\bm{a}_k^{(t)}\|_1}\sum_{\tau=0}^{\tau_k^{(t)}-1}a_{k,\tau}^{(t)}(\nabla f_k(\bm{w}_{k,t}^\tau)-\nabla\ell(\bm{w}_{k,t}^\tau))\right\|^2\right] \nonumber \\
  &\quad +2\mathbb{E}\left[\left\|\frac{1}{\|\bm{a}_k^{(t)}\|_1}\sum_{\tau=0}^{\tau_k^{(t)}-1}a_{k,\tau}^{(t)}\nabla\ell(\bm{w}_{k,t}^\tau)\right\|^2\right] \nonumber \\
  &\leq2(\sigma^2+G^2), 
  \label{eq:tmp5}
\end{align}
where the first inequality follows from $\|\bm{a}+\bm{b}\|^2\leq2\|\bm{a}\|^2+2\|\bm{b}\|^2$. 
The final inequality is derived from Jensen's inequality.
By substituting \eqref{eq:tmp5} into \eqref{eq:tmp4}, 
we obtain 
\begin{align}
  \mathcal{T}_3&\leq\frac{2^K(\sigma^2+G^2)}{T}\sum_{k=1}^K\sum_{t=0}^{T-1}(\Theta_k^{(t)}-\theta_k)^2 \nonumber \\
  &= 2^K(\sigma^2+G^2)\sum_{k=1}^K\frac{1}{T}\sum_{t=0}^{T-1}\left(\Theta_k^{(t)}-\frac{1}{T}\sum_{t'=0}^{T-1}\Theta_k^{(t')}\right)^2.
  \label{eq:tmp7}
\end{align}

Finally, the following inequality holds: 
\begin{equation}
  \min_t\mathbb{E}\left[\left\|\nabla f(\bm{w}^{(t)})\right\|^2\right] \leq \frac{1}{T}\sum_t\mathbb{E}\left[\left\|\nabla f(\bm{w}^{(t)})\right\|^2\right].
\end{equation}
Therefore, from \eqref{eq:tmp6} and \eqref{eq:tmp7}, 
we obtain the statement in Theorem~\ref{theo:convergence}.
By setting the learning rate to $\eta=\sqrt{K/(\tilde{\tau}T)}$, 
which satisfies the constraints, 
where $\bar{\tau}^{(t)}=\sum_{k=1}^K\tau_k^{(t)}/K$ and $\tilde{\tau}=\sum_{t=0}^{T-1}\bar{\tau}^{(t)}/T$, 
we then have 
\begin{align}
  &\min_t \mathbb{E}\left[\|\nabla f(\bm{w}^{(t)})\|^2\right] \nonumber \\
  &\leq\mathcal{O}\left(\frac{\tilde{\tau}}{\tau_{\mathrm{eff}}}\frac{1}{\sqrt{K\tilde{\tau}T}}\right) 
  +\mathcal{O}\left(\frac{\tilde{A}\sigma^2}{\sqrt{K\tilde{\tau}T}}\right) \nonumber \\
  & \quad+\mathcal{O}\left(\frac{K\tilde{B}\sigma^2}{\tilde{\tau}T}\right)
  +\mathcal{O}\left(\frac{K\tilde{C}\kappa^2}{\tilde{\tau}T}\right) 
  +\mathcal{O}\left(T^\delta\right), 
\end{align}
where the order due to the variance of learned weights \eqref{eq:tmp7} is summarized in $\mathcal{O}\left(T^\delta\right)$.

\begin{IEEEbiography}[{\includegraphics[width=1in,height=1.25in,clip,keepaspectratio]{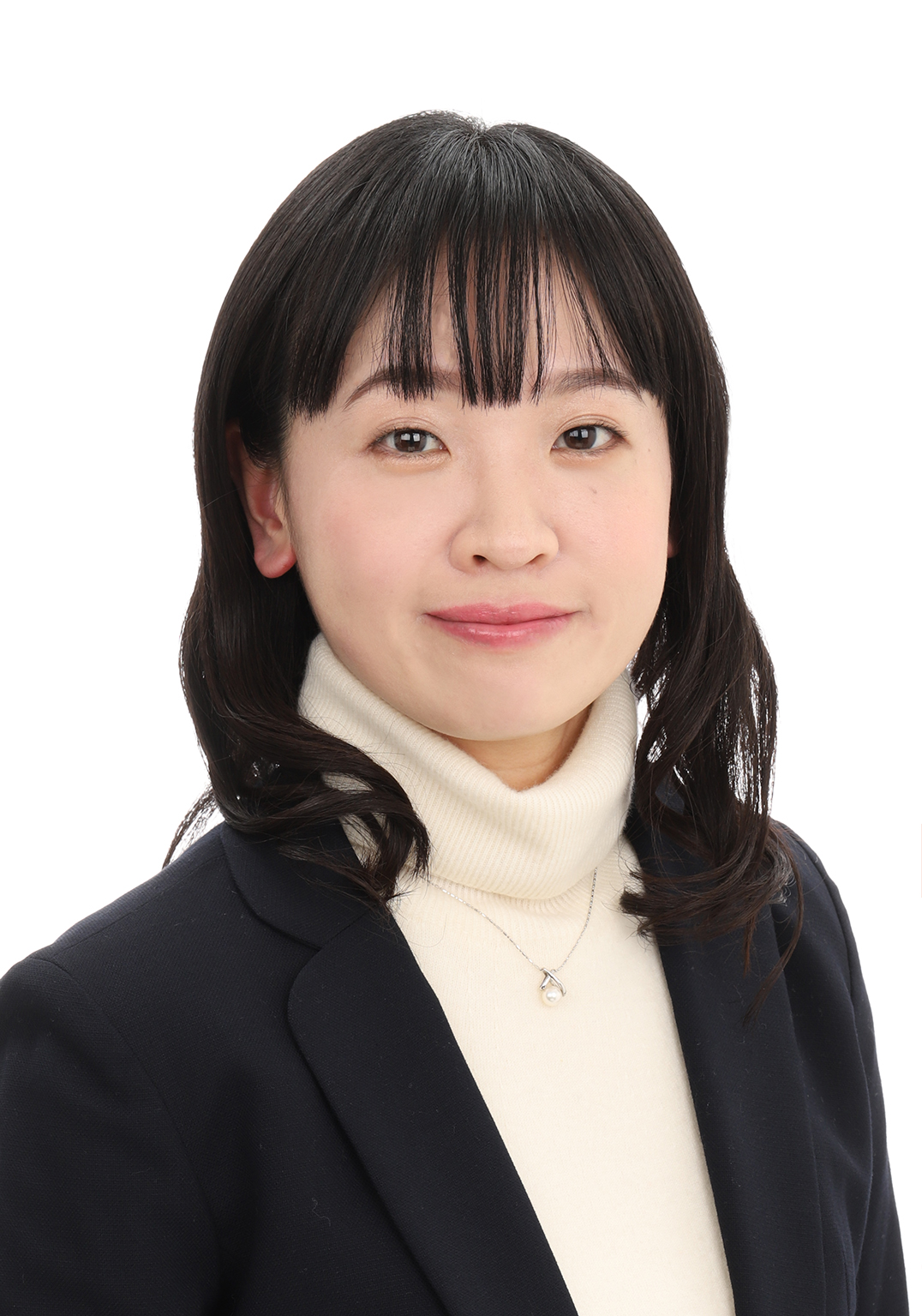}}]
{Ayano Nakai-Kasai}~(Member, IEEE)~received the bachelor’s degree 
in engineering, the master’s degree in informatics,
and Ph.D. degree in informatics from Kyoto University, Kyoto, Japan, in 2016, 2018, and 2021,
respectively. She is currently an Assistant Professor
at Graduate School of Engineering, Nagoya Institute of Technology. Her research interests include
signal processing, wireless communication, and machine learning. She received the Young Researchers’
Award from the Institute of Electronics, Information 
and Communication Engineers in 2018 and APSIPA 
ASC 2019 Best Special Session Paper Nomination Award. She is a member
of IEEE and IEICE.
\end{IEEEbiography}

\begin{IEEEbiography}[{\includegraphics[width=1in,height=1.25in,clip,keepaspectratio]{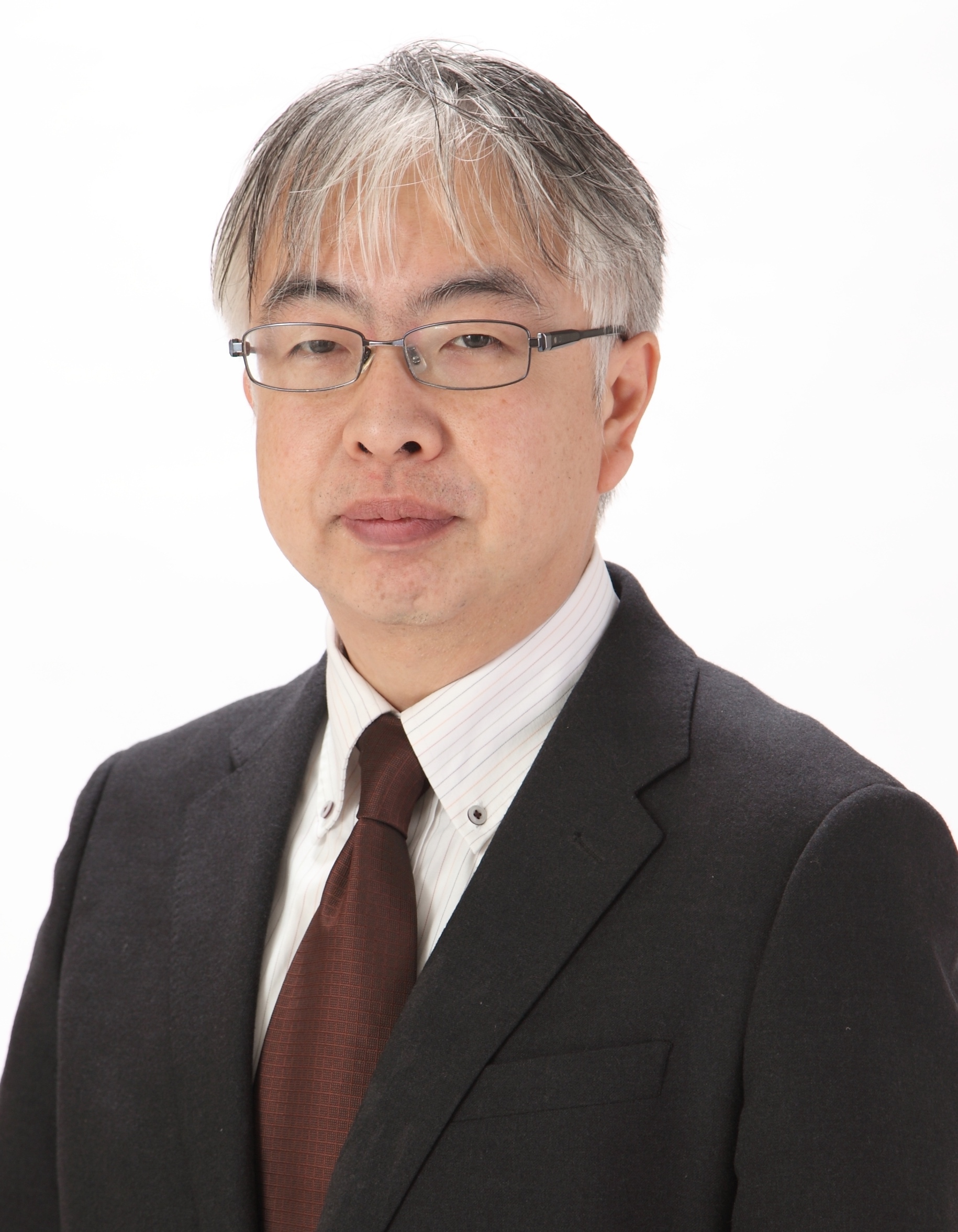}}]
{Tadashi Wadayama}~(Member, IEEE)~was born in Kyoto, Japan, on May 9,1968.
He received the B.E., the M.E., and the D.E. degrees from Kyoto Institute of Technology in 1991, 1993 and 1997, respectively.
On 1995, he started to work with Faculty of Computer Science and System Engineering, Okayama Prefectural University as a research associate.
From April 1999 to March 2000, he stayed in Institute of Experimental Mathematics, Essen University (Germany) as a visiting researcher.
On 2004, he moved to Nagoya Institute of Technology as an associate professor. Since 2010, he has been a full professor of Nagoya Institute of Technology.
His research interests are in coding theory, information theory, and signal processing for wireless communications.
He is a member of IEEE and a senior member of IEICE.
\end{IEEEbiography}



\vfill\pagebreak

\end{document}